\documentclass{article}

\usepackage[accepted]{icml2021}

\makeatletter
\newcommand{\maketitlenew}{\@maketitle}
\makeatother

\usepackage[utf8]{inputenc} 
\usepackage[T1]{fontenc}    
\usepackage{hyperref}       
\usepackage{url}            
\usepackage{booktabs}       
\usepackage{amsfonts}       
\usepackage{nicefrac}       
\usepackage{microtype}      
\usepackage{xcolor}         
\usepackage{graphicx}
\usepackage{subfigure}

\usepackage{amssymb}
\usepackage{amsmath}
\usepackage{esint}           
\usepackage{mathtools}
\usepackage{bbm}

\usepackage{wrapfig}
\usepackage{tikz}
\usetikzlibrary{bayesnet}
\usetikzlibrary{positioning}
\usetikzlibrary{arrows.meta}
\usetikzlibrary{calc}

\usetikzlibrary{positioning,fit,shapes,backgrounds,circuits.logic.US}

\usepackage[symbol]{footmisc}

\usepackage[shortlabels,inline]{enumitem}  
\newlist{inlineitemize}{enumerate*}{1}
\setlist[inlineitemize]{label=(\roman*)}

\usepackage[capitalise]{cleveref} 
\crefrangeformat{equation}{#3Eqs. (#1)#4 to #5(#2)#6}

\usepackage{todonotes}
\newboolean{todo}
\setboolean{todo}{True} 

\newcommand{\remarkInternal}[4]{\ifthenelse{\boolean{todo}}{\todo[inline, color=#2, caption={2do}, #3]{\begin{minipage}{\textwidth-4pt}\emph{Remark #1:}\\#4\end{minipage}}}{}}

\DeclareMathOperator{\tr}{\mathrm{tr}\,}

\DeclareMathOperator{\R}{\mathbb{R}}

\DeclareMathOperator{\1}{\mathbbm{1} }

\DeclareMathOperator{\E}{\mathsf{E}\,}

\DeclareMathOperator{\GamDis}{\mathrm{Gam}}
\DeclareMathOperator{\CatDis}{\mathrm{Cat}}

\DeclareMathOperator{\NDis}{\mathcal{N}\,}

\DeclareMathOperator{\DirDis}{\mathrm{Dir}}
\DeclareMathOperator{\IWDis}{\mathcal{IW}}

\DeclareMathOperator{\NIWDis}{\mathcal{NIW}\,}
\DeclareMathOperator{\MNDis}{\mathcal{MN}\,}

\usepackage[nolist]{acronym} 

\icmltitlerunning{MCMC for Continuous-Time Switching Dynamical Systems}

\begin{document}
\begin{acronym}
\acro{ssa}[SSA]{stochastic sampling algorithm}
\acro{ctmc}[CTMC]{continuous-time Markov chain}
\acroplural{ctmc}[CTMCs]{continuous-time Markov chains}
\acro{sde}[SDE]{stochastic differential equation}
\acroplural{sde}[SDEs]{stochastic differential equations}
\acro{em}[EM]{expectation maximization}
\acro{vem}[VEM]{variational expectation maximization}
\acro{slds}[SLDS]{switching linear dynamical system}
\acroplural{slds}[SLDS]{switching linear dynamical system}
\acro{pde}[PDE]{partial differential equation}
\acroplural{pde}[PDEs]{partial differential equations}
\acro{ode}[ODE]{ordinary differential equation} 
\acroplural{ode} [ODEs]{ordinary differential equations} 
\acro{sde}[SDE]{stochastic differential equation} 
\acroplural{sde} [SDEs]{stochastic differential equations}
\acro{ssde}[SSDE]{switching stochastic differential equation} 
\acroplural{ssde}[SSDEs]{switching stochastic differential equations}
\acro{mjp}[MJP]{Markov jump process}
\acroplural{mjp}[MJPs]{Markov jump processes}
\acro{gp}[GP]{Gaussian process}
\acroplural{gp}[GPs]{Gaussian processes}
\acro{gpa}[GPA]{Gaussian process approximation}
\acro{kl}[KL]{Kullback-Leibler}
\acro{fpe}[FPE]{Fokker-Planck equation}
\acro{gfpe}[HME]{hybrid master equation}
\acro{md}[MD]{molecular dynamics}
\acro{hmm}[HMM]{hidden Markov model}
\acroplural{hmm}[HMMs]{hidden Markov models}
\acro{pdf}[PDF]{probability density function}
\acro{cdf}[CDF]{cumulative distribution function}
\acro{msm}[MSM]{Markov state model}
\acroplural{msm}[MSMs]{Markov state models}
\acro{vi}[VI]{variational inference}
\acro{map}[MAP]{maximum a-posteriori}
\acro{hme}[HME]{hybrid master equation}
\acro{el}[EL]{Euler-Lagrange}
\acro{elbo}[ELBO]{evidence lower bound}
\acroplural{pdmp}[PDMPs]{piecewise deterministic Markov processes}
\acroplural{shs}[SHS]{stochastic hybrid systems}
\acro{shs}[SHS]{stochastic hybrid system}
\acro{bffb}[BFFB]{backward-forward/forward-backward}
\acro{gfp}[GFP]{green fluorescent protein}
\acro{kbe}[KBE]{Kolmogorov backward equation}
\acro{niw}[NIW]{Normal-inverse-Wishart}
\acro{mn}[MN]{Matrix-Normal}
\acro{iw}[IW]{inverse-Wishart}
\acro{rts}[RTS]{Rauch-Tung-Striebel}
\acro{rna}[RNA]{ribonucleic acid}
\end{acronym}

\twocolumn[
\icmltitle{Markov Chain Monte Carlo for Continuous-Time Switching Dynamical Systems}



\icmlsetsymbol{equal}{*}

\begin{icmlauthorlist}
\icmlauthor{Lukas Köhs}{tuda}
\icmlauthor{Bastian Alt}{tuda}
\icmlauthor{Heinz Koeppl}{tuda}
\end{icmlauthorlist}

\icmlaffiliation{tuda}{Department of Electrical Engineering and Information Technology, Technische Universität Darmstadt, Darmstadt, Germany}

\icmlcorrespondingauthor{Heinz Koeppl}{heinz.koeppl@tu-darmstadt.de}

\icmlkeywords{Machine Learning, ICML}

\vskip 0.3in
]



\printAffiliationsAndNotice{}  

\begin{abstract}
Switching dynamical systems are an expressive model class for the analysis of time-series data.
As in many fields within the natural and engineering sciences,
the systems under study typically evolve continuously in time, it is natural to consider continuous-time model formulations consisting of \aclp{ssde} governed by an underlying \acl{mjp}. Inference in these types of models is however notoriously difficult, and tractable computational schemes are rare. 
In this work, we propose a novel inference algorithm utilizing a Markov Chain Monte Carlo approach.
The presented Gibbs sampler allows to efficiently obtain samples from the exact continuous-time posterior processes.
Our framework naturally enables Bayesian parameter estimation, and we also include an estimate for the diffusion covariance, which is oftentimes assumed fixed in \aclp{sde} models.
We evaluate our framework under the modeling assumption and compare it against an existing variational inference approach.
\end{abstract}

\section{Introduction}
\label{sect:introduction}
A wide range of natural and engineered systems are naturally modeled as continuous-time stochastic processes.
The model state space depends on the system at hand; while modeling approaches often focus on either a discrete or continuous state description, a great variety of systems involve both discrete and continuous components, where a discrete, switching process influences the dynamics of some continuous quantity.
In the biological sciences in particular, one is often dealing with systems with hybrid state space structure: In neuroscience, for instance, the brain is commonly assumed to adopt different states, e.g., depending on the environment or own actions, such as \emph{eyes opened} versus \emph{eyes closed}, eliciting qualitatively different electrophysiological dynamics \cite{weng2020open}.
Likewise, in cellular biology, the state of genetic toggle switches drives continuous measurable quantities such as a protein concentrations \cite{tian2006stochastic}, and the stochastic conformational gating of ion channels \cite{bressloff2020switching} determines the ion current passing through \cite{anderson2015stochastic}.
Further examples include engineering applications, such as the safety of air traffic under potential system failures \cite{lygeros2010stochastic}; the control of the power distribution in an electrical grid in different connectivity modes \cite{stvrelec2012modeling};
and analyses of exchange rates or stock returns depending on market states in econometrics \cite{azzouzi1999modelling}.

A versatile framework to analyze systems of this kind are \acp{shs}, which have a long history in control, statistics and machine learning \cite{davis1984piecewise,hu2000towards,engell2003modelling,davis2018markov,cassandras2018stochastic}.
\acp{shs} can be defined in various ways; In this paper, we focus on fully stochastic, continuous-time systems described as \acp{ssde}, in which a discrete \ac{mjp} drives a subordinate \ac{sde} \cite{xuerongStochasticDifferentialEquations2006}.

The problem of inference has been treated extensively in both \acp{sde} and \acp{mjp}.
Classical exact results include the well-known Kalman and Wonham filters \cite{bain2008fundamentals}, as well as respective smoothing extensions, such as the \ac{rts} smoother \cite{sarkka2013bayesian,vanhandelFilteringStabilityRobustness2007}.
Also, a great variety of approximate solutions have been worked out, building both on sampling schemes \cite{doucet2000sequential,rao2013fast} and variational inference approaches \cite{opperVariationalInferenceMarkov2008,wildner2021}.

Inference frameworks for \acp{ssde} are scarce, however. 
In \acp{slds}, a discrete-time analog of \acp{ssde}, inference methods utilizing  exact sampling \cite{willskyNonparametricBayesianLearning2009,linderman2017bayesian} as well as approximate optimization techniques \cite{johnson2016composing} have been proposed in recent years.
While a variational approach to \ac{ssde} inference has been proposed recently \cite{kohs2021variational}, a framework for exact inference is lacking to the best of our knowledge.
Here, we present a Gibbs sampling scheme, allowing to efficiently sample from the exact posterior \ac{sde} and \ac{mjp} processes.
To this end, we derive the posterior process equations and present a \ac{bffb}-sweeping algorithm to deal with the intricacies of stochastic integration. 
We then combine this with a Bayesian treatment of the model parameters, for which we obtain full posterior distributions.
An implementation of the proposed framework is publicly available.\footnote[3]{A link to the implementation will be added in the camera-ready version of the manuscript.}

\section{Model}
\label{sec:background}
\subsection{Mathematical Background}
\begin{figure}
    \includegraphics[width=\columnwidth]{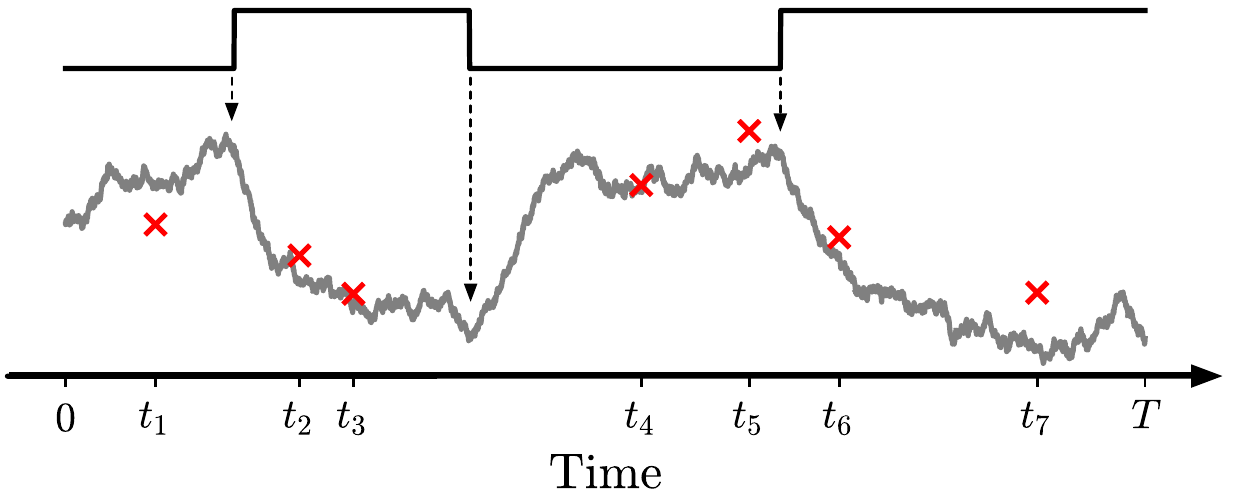}
    \caption{Sketch of a hybrid process realization. 
    A two-state \acs{mjp}  $Z(t)$ (top, see \cref{eq:transition_rates}), freely evolving in the interval $t\in[0, T]$, 
    controls the dynamics
    of the \ac{ssde} $Y(t)\mid Z(t)$ (bottom, gray line),  cf. \cref{eq:SSDE}.
    From these latent continuous dynamics, only sparse and noisy observations (red crosses) obtained at irregularly-spaced time points $t_1, t_2, \dots$ are available for inference.
    Vertical arrows indicate the
    $Z$-transitions.
}
    \label{fig:process_sketch_pgm}
\end{figure}

The switching dynamical systems we consider consist of three joint stochastic processes
\begin{inlineitemize}
\item a continuous-time switching process $Z:=\{Z(t)\}_{t\geq 0}$, 
\item a continuous-time subordinated diffusion process $Y:=\{Y(t)\}_{t \geq 0}$, and
\item an observation process $X:=\{X_i\}_{i \in \mathbb N}$ at discrete time points $\{t_i\}_{i \in \mathbb N}$.
\end{inlineitemize}
A realization of this system is shown in \cref{fig:process_sketch_pgm}. 

The \emph{switching process} $Z$, at the top of the hierarchy is given as a \acf{mjp}~\cite{norrisMarkovChains1997} freely evolving in time $t$. 
An \ac{mjp}, with $Z(t)\in \mathcal Z \subseteq \mathbb N$, is a time-continuous Markov process on a countable state space $\mathcal{Z}$ which is fully characterized by 
\begin{inlineitemize} 
\item an initial probability distribution
$p(z_0) := \mathsf P(Z(0)=z_0), \; \forall z_0 \in \mathcal Z$, and 
\item the transition rate function defined for $z' \in \mathcal Z \setminus z$ as
\end{inlineitemize}
\begin{equation}
       \Lambda(z, z',t) := \lim_{h\searrow 0}h^{-1}\mathsf P (Z(t+h)=z'\mid Z(t)=z)
    \label{eq:transition_rates}
\end{equation}
and the \emph{exit rate} $\Lambda(z, z, t) :=- \sum_{z'\in \mathcal Z \setminus z}\Lambda(z, z',t)$. 

The \emph{subordinated diffusion process} $Y$ is a continuous-valued process with $Y(t)\in \mathcal Y \subseteq \mathbb R^n$, depending on the freely evolving \ac{mjp} $Z$. This yields a \acf{ssde}~\cite{xuerongStochasticDifferentialEquations2006}
defined in an It\^{o} sense as
\begin{equation}
    \mathrm d Y(t) = f( Z(t),Y(t), t) \,\mathrm d t + Q(Z(t),Y(t), t) \,\mathrm d W(t),
    \label{eq:SSDE}
\end{equation}
with drift function $f: \mathcal Z \times \mathcal Y \times \mathbb R_{\geq 0} \rightarrow \mathcal Y$,
the invertible dispersion $Q: \mathcal Z \times \mathcal{Y} \times  \mathbb R_{\geq 0} \rightarrow \mathbb R^{n \times n}$ determining the  noise covariance as $D( z,y, t):= Q(z, y, t) Q^\top(z, y, t)$ with $y \in \mathcal Y$, $z \in \mathcal Z$.
The diffusion is driven by the $n$-dimensional standard Wiener process $W$, and the distribution of the initial value is given by the density $p(y_0 \mid z_0)$.
The difference to a conventional \ac{sde} consists in the $Z(t)$-dependence of $f$ and $Q$; for an accessible introduction to \acp{sde}, see, e.g., Särkkä \& Solin \yrcite{sarkkaAppliedStochasticDifferential2019}.
Given a realization of the \ac{mjp}, the \ac{ssde} in \cref{eq:SSDE} can hence be equivalently interpreted as a concatenation of individual \acp{sde} determined by the switching process $Z$.

The \ac{mjp} $Z$ and the diffusion $Y$ together constitute the latent \emph{hybrid process} $\lbrace Z(t), Y(t)\rbrace_{t \geq 0}$.
To avoid ambiguity, we denote the discrete value $Z(t)$ as the \emph{mode} and the continuous value $Y(t)$ as the \emph{state} of the system.
Furthermore, we will use upper-case letters $Z(t)$ to refer to random variables and lower-case letters $z(t)$ to refer to respective realizations throughout the paper.
For any time interval $[0, T]$, the hybrid process induces a measure $\mathsf P$ on the space $\Omega_T$ of all possible paths $\omega_T:= \left(z_{[0,T]}, y_{[0,T]}\right)$, where $y_{[0,T]}:= \lbrace y(t)\rbrace_{t\in [0, T]}, z_{[0,T]}:=\lbrace z(t)\rbrace_{t\in [0, T]}$~\cite{cinlar2011probability}; that is, for any event $\mathcal{A}$ in the Borel $\sigma$-algebra of paths, we can formally find its associated probability by integration,
\begin{equation}
\begin{split}
    \mathsf P\left(\left(Z_{[0,T]}, Y_{[0,T]}\right) \in \mathcal A\right) &= \int_{\mathcal A}  \mathsf P\left(\left(Z_{[0,T]}, Y_{[0,T]}\right) \in \mathrm d \omega\right) \\
    &\equiv \int_{\mathcal A}\mathrm d\mathsf P(\omega_T).
    \label{eq:path_measure_hybrid}
    \end{split}
\end{equation}

Though it is not sensible to define a density for the path-measure in \cref{eq:path_measure_hybrid} as there does not exist a Lebesgue measure for an infinite uncountable number of random variables, we note that time-point-wise this quantity admits a probability density function $p(z, y, t)$,
\begin{equation}\label{eq:expectation}
\begin{split}
    \E\left[\varphi(Z(t), Y(t), t)\right] &= \int_{\Omega}\varphi(Z(t), Y(t), t)\mathrm d\mathsf P(\omega_T)\\
    &=\sum_{z\in \mathcal Z} \int_{\mathcal Y} \varphi(z, y, t) p(z, y, t) \, \mathrm dy,
\end{split}
\end{equation}
where $\varphi: \mathcal Y \times \mathcal Z \times \R_{\geq 0} \rightarrow \R$ is an arbitrary test function.
This density evolves over time according to the \ac{gfpe}
\begin{equation}
     \partial_t p(y, z, t )     = [\mathcal{L} p](y, z, t )
      \label{eq:generalized_fkp_prior}
\end{equation}
with initial distribution $p(y_0, z_0, 0)=p(z_0) p(y_0 \mid z_0)$ and $\mathcal{L}=\mathcal{T}+\mathcal{F}$,
\begin{equation*}
    \begin{split}
               [\mathcal{T}\varphi](y,z,t) &:=\sum_{z^\prime \in \mathcal{Z}} \Lambda(z',z,t) \varphi(y,z',t)\\
             [\mathcal{F} \varphi](y,z,t) &:=- \sum_{i=1}^n \partial_{y_i} \left\lbrace f_i(y,z, t) \varphi(y,z,t)\right\rbrace  \\
             &\quad+  \frac{1}{2}\sum_{i=1}^n \sum_{j=1}^n  \partial_{y_i}  \partial_{y_j} \lbrace 
      D_{ij}(y, z, t) \varphi(y,z,t) \rbrace
    \end{split}
\end{equation*}
with $\varphi$ again an arbitrary test function, see, e.g. \cite{pawula1967generalizations,kohs2021variational}.

Unfortunately, a general, analytical solution to the \ac{gfpe} does not exist.
Numerical solution methods such as finite elements may be applied, which however---suffering from the curse of dimensionality---can only be used in very low-dimensional settings, and even in low dimensions are non-trivial to adapt to the given \ac{pde} \cite{grossmann2007numerical}.

Sampling trajectories from $\{Y(t), Z(t)\}_{t\in [0,T]}$ is straightforward, on the other hand:
A realization $z_{[0,T]}$ of the discrete process $Z$ can be simulated by utilizing the Doob-Gillespie algorithm \cite{doob1945markoff}.
Given this trajectory $z_{[0,T]}$, the diffusion $Y$ can be simulated using, e.g., an Euler-Maruyama or stochastic Runge-Kutta method~\cite{kloeden1992stochastic}.

Finally, the \emph{observation process} $X$ consists of a countable set of observed data points $\{X_i\}_{i \in \mathbb N}$, with $X_i \in \mathcal X$, at times $\{t_i\}_{i \in \mathbb N}$.
The $i$th data point $X_i$ is generated conditional on the diffusion process $Y$ as $X_i \sim p(x_i \mid Y(t_i)=y_i)$.
In general, the observation space $\mathcal{X}$ can be either discrete, $\mathcal{X}\subseteq \mathbb N^l$, or continuous, $\mathcal{X}\subseteq \R^l$.
As we provide a \emph{continuous-time} description for the latent processes while observations are generated at \emph{discrete} time points, our model belongs to the class of continuous-discrete models, on which a large body of literature exists in the field of stochastic filtering \cite{maybeck1979stochastic,daum1984,sarkkaAppliedStochasticDifferential2019}.
We emphasize that this model formulation is of great practical relevance, as data is often recorded at discrete time points, while the system of interest evolves continuously in time, see, e.g., \cite{cassandras2009introduction}.

\subsection{Modeling Assumptions}
\label{sec:model_specification}
The background presented so far was general to any hybrid processes.
For the remainder of the paper, we consider a time-homogeneous prior switching process $Z$ with rate function $\Lambda(z, z', t) = \Lambda(z, z')$ and we parametrize the initial distributions as $p(z_0) = \CatDis(z_0\mid \pi)$. 
Furthermore, we focus on mode-dependent linear time-invariant stochastic systems, i.e.,
\begin{equation}
    f(z, y, t) = f(z,y) = A(z)y + b(z),
    \label{eq:linear_drift}
\end{equation}
where the affine drift function is parameterized for each mode $z \in \mathcal Z$ by $A(z) \in \R^{n \times n}$ and $b(z)\in \R^n$. We define the shorthands $\Gamma(z):=\left[A(z), b(z)\right]\in \R^{n \times n+1}$ and $\bar{y}:= \left[y^\top, 1^\top_n\right]^\top \in \R^{n+1}$, where $1_n$ is the $n$-dimensional all-ones vector. This parametrization yields a linear system in the variable $\bar{y}$, i.e.,
$A(z)y + b(z) = \Gamma(z)\bar{y}$.
For the mode-dependent linear time-invariant stochastic system, we assume a time-homogeneous and state-independent dispersion, i.e., $Q(z, y, t) = Q(z)$
and we assume a Gaussian initial distribution, which we parameterize as $p(y_0) = \NDis(y_0\mid \mu_0, \Sigma_0)$.

Lastly, we follow a linear observation model for the data as $X_i=Y_i+\zeta$, with the zero-mean Gaussian noise $\zeta\sim \NDis(\zeta\mid 0, \Sigma_{x})$ and observation covariance $\Sigma_{x}$. Hence, the observation likelihood for the $i$th data point $X_i$ is given as
\begin{equation}
    X_i \sim \NDis(x_i\mid y_i, \Sigma_{x}).
    \label{eq:observation_dist}
\end{equation}

\section{Inference}
\label{sec:inference}
For inference, we take on a Bayesian view similar to discrete-time filtering and smoothing~\cite{sarkka2013bayesian}.
We consider a set $x_{[1,N]}:=\{x_i\}_{i=1}^N$ of $N$ observations obtained at time points $0\leq t_1, ..., t_N \leq T$.
We are interested in computing a path-wise posterior distribution over the latent hybrid process $\left\lbrace Z(t), Y(t)\right\rbrace$
on the interval $[0,T]$ and all of its parameters $\Theta$ given the observed data, i.e.,
\begin{equation}
    \mathsf P\left(\left(Z_{[0,T]}, Y_{[0,T]}, \Theta \right) \in \cdot \mid x_{[1,N]}\right).
    \label{eq:full_posterior_path_exact}
\end{equation}
As is common in Bayesian inference, computing \cref{eq:full_posterior_path_exact} is intractable as 
\begin{inlineitemize}
\item integration over a high-dimensional parameter space is hard and 
\item computing the required posterior distribution over the latent hybrid process
\end{inlineitemize}
\begin{equation*}
\begin{split}
        \mathsf P\left((Z_{[0,T]},Y_{[0,T]}) \in \mathcal A \mid x_{[1, N]}, \theta \right)\\
        =\int_{\mathcal A}\mathrm d \mathsf P(\omega_T\mid x_{[1, N]}, \theta),
\end{split}
\end{equation*}
is infeasible. This can be seen by noting that the posterior distribution over the latent hybrid process can be equivalently expressed via the \emph{smoothing distribution}, that is, the time point-wise posterior marginal density $p(z, y , t \mid x_{[1,N]}, \theta)$, cf. \cref{eq:expectation}.
For this quantity, an exact evolution equation can be derived \cite{kohs2021variational}.
While the above posterior can hence be computed in principle, this is not a feasible option even for toy systems as it requires the solution of two coupled \acp{pde} of type \labelcref{eq:generalized_fkp_prior} with discrete and continuous components.

To overcome these issues, we propose a blocked Gibbs sampler, where the switching process $Z_{[0,T]}$, the diffusion process $Y_{[0,T]}$ and the parameters $\Theta$ are sampled in turn from the complete conditional measures, i.e., the measures conditioned on each other and the data $x_{[1,N]}$:
\begin{align}
    y_{[0,T]}&\sim \mathsf P(Y_{[0,T]}\in \mathrm \cdot  \mid z_{[0,T]}, x_{[1,N]},\theta),\label{eq:Y_full_conditional}\\
    z_{[0,T]}&\sim \mathsf P(Z_{[0,T]}\in \mathrm \cdot \mid y_{[0,T]}, x_{[1,N]}, \theta),\label{eq:Z_full_conditional}\\
    \theta &\sim \mathsf P(\Theta \in \mathrm \cdot  \mid z_{[0,T]}, y_{[0,T]}, x_{[1,N]})\label{eq:T_full_conditional}.
\end{align}
Hence, this scheme yields samples from the desired joint posterior distribution in \cref{eq:full_posterior_path_exact}.

The path-wise quantities \cref{eq:Y_full_conditional,eq:Z_full_conditional} can be show to each describe conditional Markov processes.
In principle, we desire to sample from these processes in a manner akin to the forward- and backward-recursion in traditional discrete-time \acp{hmm} \cite{barber2012}.
In contrast to the discrete-time case, however, we can not obtain such recursions by an application of Bayes' rule, as it is not possible to define sensible probability densities on the path-space $\Omega$, cf. \cref{sec:background}.
Sampling from the full conditional measures is hence non-trivial.
Drawing on results from filtering and smoothing theory, we derive in the following the  evolution equations for \cref{eq:Y_full_conditional,eq:Z_full_conditional} on the process level, allowing us to generate the desired samples.
As \cref{eq:Y_full_conditional} and \cref{eq:Z_full_conditional} both depend on the same Brownian motion instance $W$, complications arise due to the asymmetry of the It\^o integral which are not found in discrete time.
To circumvent these issues, all stochastic integrations are carried out forward in time.
This results in a \acf{bffb} scheme, where the time direction of the actual path simulation is reversed between the posterior $Z$- and $Y$-paths.

We additionally sample the model parameters from the full conditional distribution \cref{eq:T_full_conditional}.
By the use of conjugate prior distributions, \cref{eq:T_full_conditional} yields closed-form distributions for all parameters.
For the dispersion $Q$, we do not obtain the posterior directly, but utilize a Metropolis-adapted Langevin scheme, ensuring numerical stability \cite{besag1995bayesian}.

We emphasize that inference schemes for this versatile class of processes are rare: to the best of our knowledge, the only framework available is a recent variational approach \cite{kohs2021variational}, which however needs to make strong approximating assumptions and does not provide Bayesian parameter estimates.
In this approach, the exact smoothing density $p(z, y, t\mid x_{[1,N]})$ is approximated by a simple mixture of \acp{gp}, which is unable to resolve non-stationary diffusion dynamics.

\subsection{Gibbs Step: Sampling the Conditional Diffusion Process $Y_{[0,T]}$}
To sample from the full conditional diffusion path measure
\begin{equation}
    y_{[0,T]} \sim \mathsf P(Y_{[0,T]}\in \mathrm \cdot\mid z_{[0,T]}, x_{[1,N]}, \theta), 
    \label{eq:full_conditional_diffusion_path}
\end{equation}
we first acknowledge that by conditioning on the process $z_{[0,T]}$ the generative model presented in \cref{sec:background} reduces to a \ac{ssde} with  a fixed switching path.
We can interpret this \ac{ssde} \labelcref{eq:SSDE} as a temporal sequence of individual \acp{sde}, or, equivalently, as we assume all drift functions \cref{eq:linear_drift} to be linear, as one \ac{sde} with an explicit time-dependence of the parameters; more concretely, we have a drift function
\begin{equation}
     f(z(t),y) = A(z(t))y+b(z(t)) \equiv f(y, t)
\end{equation}
and the dispersion $Q(z(t),y,t)=Q(z(t))\equiv Q(t)$.
Hence, the path measure in \cref{eq:full_conditional_diffusion_path} is governed by the solution of an \ac{sde} process conditioned on the data $x_{[1,N]}$. 
For a conventional, i.e., non-switching, \ac{sde},
\begin{equation}
    \mathrm dY(t) = f(Y(t), t)\mathrm dt + Q(t)\mathrm{d}W(t),
\end{equation}
it is well known that the posterior process conditioned on some data $x_{[1,N]}$ can in turn via Doob's h-transform \cite{doob1984classical,rogers2000diffusions} be expressed as an \ac{sde} with a modified drift function $\tilde{f}(Y(t), t)$:
\begin{equation}
\begin{split}
        \mathrm d Y(t) &= \tilde{f}(Y(t), t)\mathrm dt + Q(t)\mathrm d W(t),\label{eq:posterior_sde}\\
        \tilde{f}(Y(t), t)&=f(Y(t), t) + D(t) \partial_y \log \beta(Y(t), t).
\end{split}
\end{equation}
The reverse-filtered likelihood, an analogue to the backward messages in discrete-time \acp{hmm},
\begin{equation}\label{eq:reverse_filtered_likelihood}
    \beta(y, t) := p(x_{[k, N]}\mid y, t),\,t_k \geq t,
\end{equation}
has to fulfill the \ac{kbe} starting at the end-point $t=T$ with $\beta(y, T) = 1$,~\cite{sarkka2013bayesian}
\begin{multline}\label{eq:kbe}
   \partial_t \beta(y, t) = -\sum_{i=1}^n f_i(y, t) \partial_{y_i} \beta(y, t) \\ -\frac{1}{2}\sum_{i=1}^n \sum_{j=1}^n  D_{ij}(t) \partial_{y_i}  \partial_{y_j} 
     \beta(y, t).  
\end{multline}

Under the linearity assumption, the \ac{kbe} \labelcref{eq:kbe} can be evaluated in closed form, yielding a backward Kalman-type filter.
Note that these results are known, e.g., from results on smoothing for nonlinear diffusions \cite{mider2021}.
We provide the derivations in \cref{sec:app_backward_kalman} for completeness. In between observations, we find
\begin{equation}\label{eq:rho}
    \beta(y, t)=\NDis\left( x_k, ..., x_N \mid F(t) y  + m(t) , \Sigma(t) \right),\,t_k \geq t,
\end{equation}
where $F(t) \in \R^{(N-k)n\times n}$, $m(t) \in \R^{(N-k)n}$, and $\Sigma(t) \in \R^{(N-k)n\times (N-k)n}$ are determined by a set of \acp{ode}.
Note, however, that the support of this distribution increases with each observation incorporated.
By re-interpreting the Gaussians as distributions over $y$ rather than $x$, this can be decomposed as 
\begin{equation}\label{eq:log_rho}
    \log \beta(y, t) = -c(t) - \frac{1}{2}y^\top I(t) y + a(t)^\top y,
\end{equation}
where $c(t), I(t)$ and $a(t)$ depend on the parameters of \cref{eq:rho}.
Importantly,
\begin{inlineitemize}
\item these parameters are fixed in size, $c(t)\in \R$, $I(t)\in\R^{n\times n}$, and $a(t)\in \R^n$; and
\item we do not require the normalizer $c(t)$, as \cref{eq:posterior_sde} only depends on the gradient
\end{inlineitemize}
\begin{equation}
    \partial_y\log \beta(y, t) = -I(t)y + a(t).
\end{equation}
For these parameters, the \ac{kbe} yields a continuous-time backward analogue to the discrete-time information filter \cite{stengel1994},
\begin{equation}\label{eq:information_filter}
\begin{split}
    \frac{\mathrm d }{\mathrm d t} I(t)&=-A(t)^\top I(t) - I(t)A(t)+ I(t)D(t)I(t), \\
    \frac{\mathrm d }{\mathrm d t} a(t)&=-A(t)^\top a(t) + I(t)D(t)a(t) + I(t)b(t).
    \end{split}
\end{equation}
At the observation times, the \ac{ode} solutions are subject to the usual reset conditions~\cite{kushner1964differential}
\begin{equation}
       I(t_i) = \Sigma_{x}^{-1} + I(t_i^+),\quad a(t_i) = \Sigma_{x}^{-1}x_i + a(t_i^+),
\end{equation}
where $a(t_i^+) := \lim_{h\searrow 0}a(t_i + h)$.
Having computed $\partial_y \log \beta(y, t)$ backward from $t=T$ to $t=0$, we can then straightforwardly simulate the \ac{sde} \labelcref{eq:posterior_sde} forward in time. Hence, we can simulate the \ac{sde} 
\begin{equation}
\begin{split}
        \mathrm d Y(t) = \left\{\left[A(z(t))-Q(z(t))Q^\top(z(t))I(t)\right]Y(t)\right.\\
        \left.+ b(z(t)) +a(t)\right\} \, \mathrm d t + Q(z(t))\,\mathrm d W(t),
\end{split}
\label{eq:diffusion_full_conditional_sde_linear}
\end{equation}
which yields samples from the full conditional distribution in \cref{eq:full_conditional_diffusion_path}.
For the initial value of the full-conditional path measure, we have
\begin{equation*}
    y(0) \mid  z_{[0,T]}, x_{[1,N]}, \theta \sim \NDis\!\left(y_0\mid \bar{\mu}, \bar{\Sigma}\right),
\end{equation*}
where we show in \cref{sec:app_posterior_ssde_init} that
\begin{equation}
    \bar{\mu} = \bar{\Sigma} (\Sigma_0^{-1}\mu_0 + a(0)),\quad \bar{\Sigma} = \left(\Sigma_0^{-1} + I(0)\right)^{-1}.
\end{equation}

\subsection{Gibbs Step: Sampling the Conditional Switching Process $Z_{[0,T]}$}
With the simulated \ac{ssde} path $y_{[0,T]}$, we aim to sample from the switching full conditional path measure,
\begin{equation*}
    z_{[0,T]}\sim \mathsf P(Z_{[0,T]}\in \mathrm \cdot \mid y_{[0,T]}, x_{[1,N]}, \theta),
\end{equation*}
which reduces due to the Markovian structure described in \cref{sec:background} to
\begin{equation}\label{eq:conditional_Z_measure}
    z_{[0,T]} \sim \mathsf P(Z_{[0,T]}\in \mathrm \cdot\mid y_{[0,T]}, \theta).
\end{equation}
In principle, we would like to pursue a similar approach as in the preceding section, i.e., a backward-filtering, forward-sampling scheme.
Appropriate theoretical results exist for pure \ac{sde} systems with state-independent drift \cite{pardouxt1980stochastic,vanhandelFilteringStabilityRobustness2007}.
The state-dependence of the \ac{ssde} drift function \cref{eq:SSDE} does however not admit an \ac{ode} formulation of the reverse-filtered likelihood analogous to \cref{eq:reverse_filtered_likelihood} \cite{crisan2013robust,davis1979pathwise}.
Instead, this quantity will depend on a stochastic integral with respect to the same Brownian motion $W$ that generated the forward diffusion process.
Roughly speaking, when trying to obtain the sought-after reverse quantity starting from $t=T$, one would have to integrate with respect to a process that is known for every $t'<t$ and hence ``looks into the future''.
It is not obvious whether it is possible to find such a reverse process, but see, e.g., \cite{nualart1988stochastic,nualart2006malliavin} on anticipative stochastic calculus.
We hence adopt a reversed approach, where we first compute the filtering distribution forward in time and subsequently simulate backwards, as we then can jointly solve the occurring stochastic integrals in the same time direction, circumventing the above issues.

We first compute the filtering densities $p_f(z, t)$, which follow from the conditional path measure $\mathsf P(Z_{[0,t]}\in \mathrm dz_{[0,t]}\mid y_{[0,t]})$ as an expectation:
\begin{align}
    p_f(z, t) &= \E\left[\mathbbm 1(z(t) = z)\mid y_{[0,t]}\right]\nonumber\\
            &= \int \mathbbm 1(z(t) = z)\mathsf P(Z_{[0,t]}\in \mathrm dz_{[0,t]}\mid y_{[0,t]}).
            \label{eq:filter_def_z_process}
\end{align}
This path measure is expressible via the measure of a standard Brownian motion $W$, $\mathsf P(W_{[0,t]}\in \mathrm dy_{[0,t]})$, utilizing Girsanov's theorem~\cite{oksendal2003stochastic}
\begin{multline}
\mathsf P\left(Z_{[0,t]}\in \mathrm dz_{[0,t]} \mid y_{[0,t]}\right)\propto G(z_{[0,t]}, y_{[0,t]})\cdot\\
\mathsf P(W_{[0,t]}\in \mathrm dy_{[0,t]})\mathsf P\!\left(Z_{[0,t]}\in \mathrm dz_{[0,t]}\right)
\end{multline}
with the Radon-Nikodym derivative between the conditional and the Brownian motion measure,
\begin{align}\label{eq:radon-nikodym}
    &G(z_{[0,t]}, y_{[0,t]}) := \frac{\mathrm d \mathsf P(Y_{[0,t]}\in \mathrm dy_{[0,t]}\mid z_{[0,t]})}{\mathrm d \mathsf P(W_{[0,t]}\in \mathrm dy_{[0,t]})}\\
    &{}=\exp\left\lbrace\int_0^t f^\top(z(s), y(s))D^{-1}(z(s))\mathrm dy(s)\right.\nonumber \\
&\quad\left.- \frac{1}{2}\int_0^t f^\top(z(s), y(s))D^{-1}(z(s))f(z(s), y(s))\mathrm ds\right\rbrace.\nonumber
\end{align}

By computing a stochastic differential equation for \cref{eq:filter_def_z_process} utilizing  It\^{o}'s lemma
yields a Kushner-Stratonovich \ac{sde} describing its time evolution \cite{del2017stochastic}
\begin{equation}\label{eq:kushner-stratonovich}
\begin{split}
&\mathrm d p_f(z, t) = \sum_{z'\in \mathcal Z} \Lambda(z', z, t)p_f(z', t)\mathrm dt + p_f(z, t)\cdot\\
&{}(f(y, z) -\bar{f}(y, t))^\top D^{-1}(z)(\mathrm dy(t) - \bar{f}(y, t)\mathrm dt)
\end{split}
\end{equation}
with $\bar{f}(y, t) = \sum_{z'\in\mathcal Z}f(z', y)p_f(z', t)$; for a detailed mathematical derivation, see \cref{sec:app_kushner-stratonovich}.
Note that for $f(y, z) = f(z)$, we recover the classical Wonham filter \cite{wonham1964some}.

It is a known result that the smoothing density $p_s(z, t) := p(z, t\mid y_{[0, T]})$ admitted by the sought-after switching path measure in \cref{eq:conditional_Z_measure} can be expressed as a backward master equation via the forward filtered density $p_f(z, t) := p(z, t \mid y_{[0,t]})$~\cite{andersonSmoothingAlgorithmsNonlinear1983,vanhandelFilteringStabilityRobustness2007}. 
The backward master equation reads
\begin{equation}
    \frac{\mathrm d}{\mathrm d t}p_s(z, t) = -\sum_{z'\in \mathcal{Z}} \tilde{\Lambda}(z', z, t)p_s(z', t),
    \label{eq:z_switching_smoothing_master}
\end{equation}
with the end-point condition $p_s(z, T) = p_f(z, T)$. 
The backward rates in \cref{eq:z_switching_smoothing_master} are defined as
\begin{equation*}
        \tilde{\Lambda}(z, z', t) := \lim_{h\searrow 0}h^{-1}\mathsf P(Z(t-h)=z'\mid Z(t)=z),
\end{equation*}
and determined by the filtering density,
\begin{equation}\label{eq:backward_rates}
\tilde{\Lambda}(z', z, t) = \frac{p_f(z', t)}{p_f(z, t)}\Lambda(z, z'),
\end{equation}
with the usual exit rates $\tilde{\Lambda}(z, z, t) := -\sum_{z'\neq z}\tilde{\Lambda}(z, z', t)$.

This allows us to backward-sample a new path $z_{[0,T]}$ after forward-filtering via \cref{eq:kushner-stratonovich}. 
To simulate from the conditional switching process $Z$ with time-dependent rates \cref{eq:backward_rates}, we utilize the thinning algorithm, which generalizes the standard Doob-Gillespie algorithm \cite{lewis1979simulation}.

\begin{algorithm}
   \caption{\ac{bffb} Gibbs Sampling for Continuous-Time Switching Dynamical Systems}
   \label{alg:sampler}
\begin{algorithmic}[1]
   \STATE {\bfseries Input:} observation data $\{t_i, x_i\}_{i=1, ..., N}$
   \STATE Initialize $z^0_{[0,T]}, y^0_{[0,T]}, \theta^0$
   \FOR{i = 0, ..., NumSamples}
    \STATE Given $z^i_{[0,T]}$, compute $\partial_y \log\beta$ using \cref{eq:information_filter} 
    \STATE Given $z^i_{[0,T]}$, sample $y^{i+1}_{[0,T]}$ according to \labelcref{eq:diffusion_full_conditional_sde_linear}
    \STATE Given $y^{i+1}_{[0,T]}$, compute $p_f$ using \cref{eq:kushner-stratonovich}
    \STATE Given $y^{i+1}_{[0,T]}$, sample $z^{i+1}_{[0,T]}$ according to \cref{eq:backward_rates}
    \STATE Given $z^{i+1}_{[0,T]}, y^{i+1}_{[0,T]}$, sample model parameters $\theta^{i+1}$
   \ENDFOR
\end{algorithmic}
\end{algorithm}

\subsection{Gibbs Step: Sampling the Parameters $\Theta$}
\label{sec:gibbs_parameters}
Our framework naturally lends itself to Bayesian parameter estimation.
In the following, we specify the used prior distributions over the model parameters
and provide the resulting full conditionals one at a time.
With this model definition, the prior parameter distributions are conjugate to the respective likelihoods.
The posterior distributions hence are found by updating the distribution hyperparameters.
Due to space constraints, we omit the densities of the used distributions as well as the mathematical details of the updates and state here only the results, but provide all definitions and derivations in \cref{sec:app_parameter_updates}.
For a comprehensive overview over (conjugate) distributions, see, e.g., \cite{gelman2013bayesian}.

\paragraph{\ac{mjp} Initial Conditions}
We impose a Dirichlet prior with hyper-parameter $\alpha_{z_0}$ on the initial \ac{mjp} state distribution parameter $\pi_{z_0}$, resulting in
\begin{align}
   p(\pi_{z_0} \mid z_{[0,T]}) 
   &= \DirDis(\pi_{z_0}\mid \alpha_{z_0}+ \delta_{z(0)}),
\end{align}
with the point mass $\delta_{z(0)}$ on the value $z(0)$. 
Note that we suppress all variables in the conditioning set that $\pi_{z_0}$ is conditionally independent of.
We shall keep with this notation in all following update equations.

On the \ac{ssde} initial distribution parameters $\mu_0, \Sigma_0$ we place a \ac{niw} prior, with hyper-parameters $(\eta,\lambda,\Psi,\kappa)$, yielding
\begin{align}
    p(\mu_0, \Sigma_0\mid y_{[0,T]})= \NIWDis\!\left(\mu_0, \Sigma_0\,\middle|\,\tilde{\eta}, \tilde{\lambda}, \tilde{\Psi}, \tilde{\kappa}\right)
\end{align}
with 
\begin{equation}
    \begin{split}
        &\tilde{\eta} = \frac{\lambda\eta+y_0}{\lambda+1},\quad\,\, \tilde{\lambda} = \lambda+1,\quad\,\,  \tilde{\kappa} = \kappa+1,\\
        &\tilde{\Psi} = \left(\Psi^{-1}+\lambda\tilde{\lambda}^{-1}(y_0-\eta)(y_0-\eta)^\top \right)^{-1}.\\
    \end{split}
\end{equation}

\paragraph{\ac{mjp} Rates}
We assume the prior rates to be given by a Gamma distribution: Introducing the shorthand $\Lambda_{zz'}:= \Lambda(z, z')$,
\begin{equation}
    p\left(\Lambda_{zz'}\right) = \GamDis(\Lambda_{zz'}\mid s, r)
\end{equation}
with the shape $s\in \R_{>0}$ and rate parameter $r\in\R_{>0}$.
Characterizing the simulated path $z_{[0,T]}$ via
\begin{inlineitemize}
\item the \emph{sojourn times} $\lbrace\tau_k\rbrace$ between jumps, and
\item the \emph{state sequence} $\lbrace z_k\rbrace$, $k=0,...,K$,
\end{inlineitemize}
allows to express the path likelihood $p(z_{[0,T]}\mid \Lambda_{zz'})$ via the observed transitions $N_{zz'}=\sum_{k=0}^{K-1} \1(z_k=z \land z_{k+1}=z')$ and cumulative sojourn times $T_z=\sum_{k=0}^K \1(z_k=z)\tau_k$, 
yielding a posterior Gamma distribution for the rates,
\begin{flalign}
p(\Lambda_{z z'} \mid z_{[0,T]}) &= \GamDis(\Lambda_{z z'}\! \mid\! s +N_{zz'},r+T_z ).
\end{flalign}

\begin{figure*}[t!]
\includegraphics[width=\textwidth]{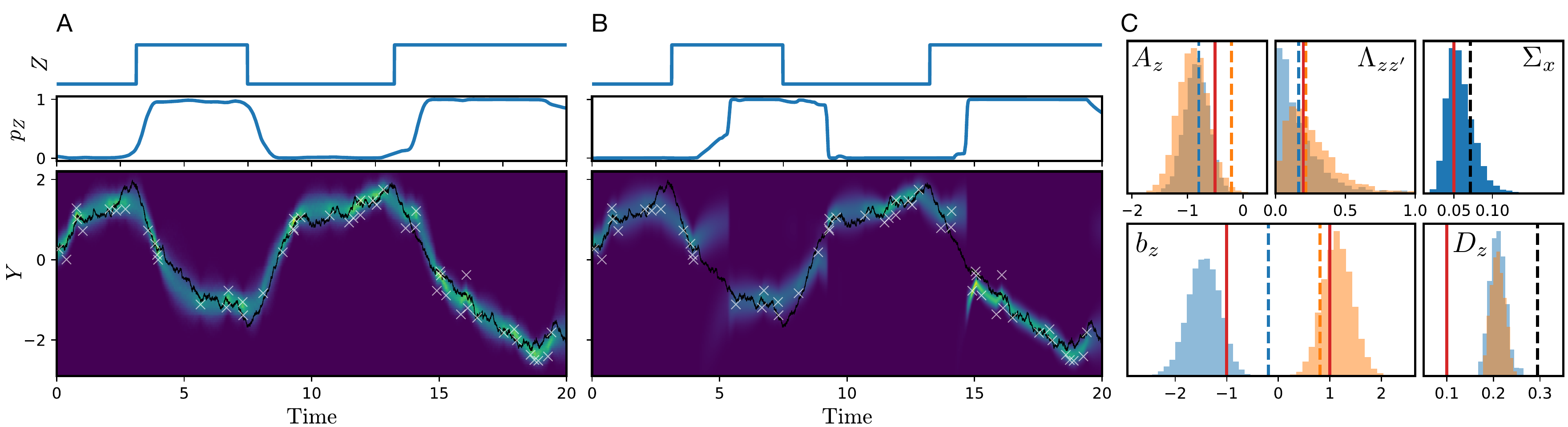}
\caption{Model validation on synthetic data and comparison with variational results. A: Results of method. Top: ground-truth switching trajectory $z_{[0,T]}$. Middle: empirical posterior $p(z, t\mid x_{[1,N]})$. Brighter colors indicate higher probability density. Bottom: respective posterior $p(y, t\mid x_{[1,N]}$. Black solid line: ground-truth latent trajectory $y_{[0,T]}$. White crosses: observations. $N_{\mathrm{samples}} = 10000$.
    B: Results of the variational method \cite{kohs2021variational}. Middle: variational posterior $q(z, t\mid x_{[1,N]})$. Bottom: variational posterior $q(y, t\mid x_{[1,N]})$.
    C: Parameter estimates of the drift parameters $A(z), b(z)$, cf. \cref{eq:linear_drift}, the \ac{mjp} rates $\Lambda(z, z')$, the \ac{sde} covariance $D(z)$ and the observation covariance $\Sigma_{x}$. Red lines: true values. Blue and orange shading indicates the two modes $z=1, 2$ where applicable. Dashed lines: variational point estimates.}
    \label{fig:1D_synSSDE}
\end{figure*}

\paragraph{\ac{sde} Drift Parameters}
In the following, we utilize as above the shorthand $\Gamma_z:=\Gamma(z)$.
The \ac{ssde} parameters $\Gamma_z$ are specified via a \ac{mn} prior
\begin{equation}
    p(\Gamma_z) = \MNDis\!\left(\Gamma_z\mid M_z, D_z, K_z\right),
\end{equation}
where $D_z=D(z)$ is the \ac{ssde} covariance.
Expressing the conditional $Y$-posterior via the Radon-Nikodym derivative $G(z_{[0,T]}, y_{[0, T]})$, c.f. \cref{eq:radon-nikodym}, we can interpret $G$ as the likelihood of the drift parameters, $G(z_{[0,T]}, y_{[0, T]}) = G(z_{[0,T]}, y_{[0, T]} \mid \lbrace\Gamma_z\rbrace)$, 
\begin{flalign}\label{eq:posterior_drift_parameters}
    p(\Gamma_z\!\mid\! z_{[0, T]}, y_{[0,T]}) \propto G(z_{[0,T]}, y_{[0, T]} \!\mid\! \lbrace\Gamma_z\rbrace)p(\Gamma_z).
\end{flalign}
We show in \cref{sec:app_parameter_updates} that the prior is conjugate with respect to $G$ when $y_{[0,T]}$ is simulated utilizing an Euler-Maruyama solver.
Accordingly,
\begin{flalign}
    p(\Gamma_z\!\mid\!y_{[0, T]}, z_{[0,T]}) &= \MNDis\!\!\left(\Gamma_z \!\mid\! \tilde{M}_z, D_z, \tilde{K}_z\right).
\end{flalign}
See \label{sec:app_parameter_updates} for the hyperparameter update equations.
Note that this prior does not guarantee stability of the individual modes, as it does not impose any constraints on the eigenvalue spectrum of the sub-matrices $A(z)$.
However, it is known that for switching systems, global stability of the system does not require strict intra-mode stability \cite{mao1999stability}. 
Conditioned on data from a stable mode, the posterior will in any event be likely peaked around stable matrices.\looseness-1

\paragraph{\ac{sde} Dispersion}
The dispersion $Q(z)$ has a special role among the model parameters, as together with the used time step-size, it determines the accuracy of the \ac{sde} solver.
While a posterior dispersion can be derived in the same vein as for the drift parameters \cref{eq:posterior_drift_parameters}, the resulting posteriors may cause the solver to become unstable if $Q$ becomes too large.
One may approach this issue by utilizing adaptive step-size solvers \cite{kloeden1992stochastic}.
For simplicity, we instead apply a Metropolis-adapted Langevin sampling scheme \cite{roberts1998optimal} for $Q(z)=Q_z$, applying the usual shorthand.
We simulate an \ac{sde} on the space of dispersion matrices with step-size $0<\xi\ll1$ and $\varepsilon\sim \NDis(0, \mathbb I_n)$~\cite{roberts1998optimal},
\begin{equation}\label{eq:mala_sde}
 Q_z^* = Q_z + \xi \partial_{Q_z}\log p(Q_z \mid y_0, y_{h}, \ldots, y_{Lh}) + \sqrt{2\kappa}\varepsilon.
\end{equation}
Here, $p(Q_z\mid \lbrace y_{hl}\rbrace)$ is the approximation of $p(Q_z\mid y_{[0,T]}) \propto G(z_{[0,T]}, y_{[0,T]})p(Q_z)$ 
on the \ac{sde} simulation time grid, $t_0=0, t_1=h,...,t_L=Lh=T$ with time-step $h$.
In practice, we parameterize $Q_z$ via
\begin{equation}
    D_z = Q_zQ_z^\top \sim \IWDis(D_z\mid \Psi_{D_z}, \lambda_{D_z}).
\end{equation}
Note that as shown in \cref{sec:app_parameter_updates}, the approximate density $p(Q_z\mid \lbrace y_{hl}\rbrace)$ is equivalent to a product of $L-1$ Gaussian transition distributions $\NDis(y_l\mid y_{l-1}, D_z h)$.
We then utilize a Metropolis rejection scheme with acceptance probability
\begin{equation}
    A(Q, Q^*) = \frac{p(Q_z^*\mid y_{[0,T]})q(Q\mid Q^*)}{p(Q_z\mid y_{[0,T]})q(Q^*\mid Q)},
\end{equation}
where $q$ denotes the (Gaussian) proposal density induced by \cref{eq:mala_sde}.
Note that we expect slow mixing of the sampler with respect to $Q$, which is due to the fact that the measures of two diffusion processes with different dispersions are singular with respect to each other \cite{shephard1997likelihood}.
As potential solutions to this peculiar issue are quite involved, an in-depth analysis of this issue is outside the scope of the present study, but see , e.g., \cite{shephard1997likelihood,golightly2008bayesian}.
For our purposes, we are satisfied with a tractable and numerically stable posterior and hence defer respective extensions to future works.

\paragraph{Observation Covariance}
Lastly, we impose an \ac{iw} prior on the observation covariance $\Sigma_{x}$, resulting in $p(\Sigma_{x}\mid x_{[1,N]}) = \IWDis(\Sigma_{x}\mid \tilde{\Psi}_{x}, \tilde{\lambda}_{x})$.
We summarize the full Gibbs sampling algorithm in \cref{alg:sampler}.\looseness-1

\section{Results}
We first verify the method on data generated under the modeling assumption and compare with the existing variational framework \cite{kohs2021variational}.
Subsequently, we use the Gibbs sampler to infer the latent expression states of an inducible gene expression system.
In all experiments, the hyperparameters are set empirically.
For other options, see, e.g. \cite{casella2001empirical}.
We initialize the Gibbs sampler in the same way to start at reasonable parameter values such as to achieve fast burn-in.
All hyperparameters are provided in \cref{sec:app_experimental_details}.
\label{sec:results}
\subsection{Verification on Ground-Truth Data}
We test the method on synthetic data from a two-mode, one-dimensional switching dynamical system as specified in \cref{sec:model_specification}.
Our Gibbs sampling scheme is able to faithfully recover the latent ground-truth trajectories; both $z_{[0,T]}$ and $y_{[0,T]}$ are reproduced with high fidelity, see \cref{fig:1D_synSSDE}.
In particular, we note that the smooth relaxation from one set-point to the other upon a switch of the $Z$-process is accurately reconstructed.
For comparison, we run the variational method \cite{kohs2021variational} on the same observations $x_{[1,N]}$.
This method returns approximate posterior marginal densities $q(z, t\mid x_{[1,N]})$ and $q(y, t\mid x_{[1,N]})$.
The variational framework fails to capture the non-stationary transition periods upon $Z$-switches and results in bimodal posterior marginals, exhibiting a ``gap'' in the posterior density $q(z, t\mid x_{[1,N]})$, which is also reflected in delayed mode transitions.
Furthermore, we obtain accurate Bayesian parameter estimates.
Note that all posteriors except the diffusion covariance $D_z$ cover the ground truth very well.
While the distribution of the latter is in total closer to the ground truth than the variational estimate, we observe slow mixing in this parameter, as discussed in \cref{sec:gibbs_parameters}.

\begin{figure}
    \centering
    \includegraphics[width=\columnwidth]{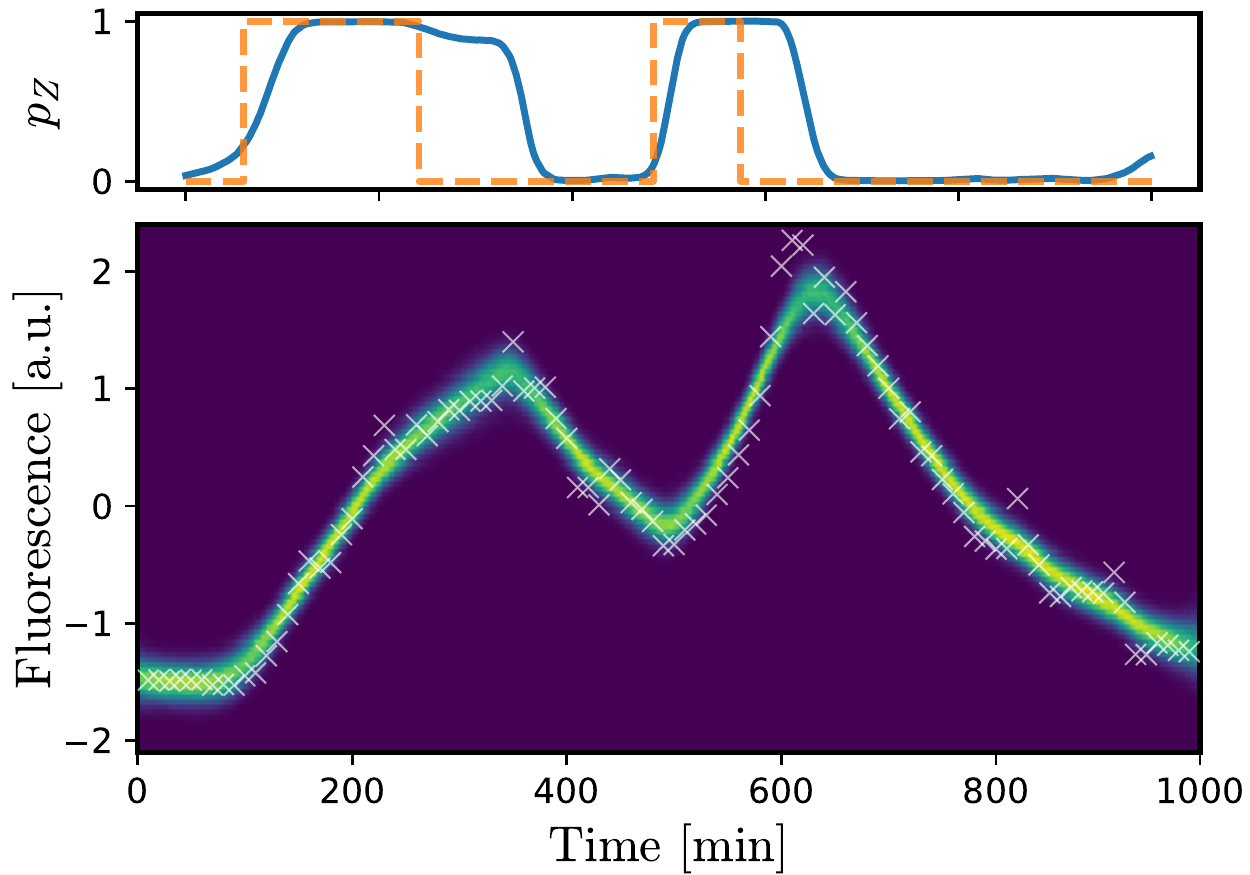}
    \caption{Inference of promoter states for an inducible gene expression system. Top: empirical posterior $p(z, t\mid x_{[1,N]})$ (blue) and chemical control (orange; ``off'', state 1, and ``on'', state 2). Note that this does not directly correspond to the promoter-``on'' and promoter-``off'' state, as a the inducer has to diffuse in and out the cell and its nucleus. Bottom: empirical posterior $p(y, t\mid x_{[1,N]})$. White crosses: observed data. $N_{\mathrm{samples}}=10000.$}
    \label{fig:gene_switching}
\end{figure}
\subsection{Inference of Gene-Switching Dynamics}
We use our framework to infer the switching dynamics of an inducible gene system measured in-house. We expressed an inducible \ac{gfp} in the eucaryotic model organism \emph{Saccharomyces cerevisiae}.
Using a microfluidic platform, gene expression can be induced at arbitrary time points by a chemical control signal utilizing $\beta$-estradiol \cite{hofmann2019tightly}.
Upon induction, expression of the \ac{gfp}-encoding gene is initiated.
We measure the amount of \ac{gfp} fluorescence over time using fluorescence microscopy.
The dynamics of transcription and translation is commonly modeled by switching \ac{sde}s with rate parameters depending on the stochastic promoter state of the gene (``on'' vs. ``off'') \cite{ocone2013hybrid}.
We aim to infer the latent stochastic promoter state, the \ac{gfp} level and the rate parameters from a set of noisy microscopy measurements. Although we have no ground truth available, we can rationalize the inferred promoter activity shown in the upper panel of Fig.~\ref{fig:gene_switching} in the context of the available inducer control signal. Upon addition of $\beta$-estradiol to the medium, a certain delay is incurred through molecular diffusion until the promoter gets activated. Similarly, upon removal of the chemical from the medium, promoter deactivation is governed by the export of the chemical from the cell through diffusion. Along the same lines, the inferred \ac{gfp} level shown in the lower panel of Fig.~\ref{fig:gene_switching} is in line with the elementary processes of \ac{rna} transcription and protein translation that result in a delayed activation and deactivation on the protein level with respect to the promoter state. The  estimates of the rate parameters are provided in \cref{sec:app_experimental_details}.

\section{Discussion}
\label{sec:discussion}
We presented, to the best of our knowledge, the first tractable sampling-based path-space inference scheme for discretely observed continuous-time switching dynamical systems.
This enables accurate Bayesian posterior inference in settings where existing variational inference methods~\cite{kohs2021variational} fail.
We derived a blocked Gibbs sampler, where we generate sample paths from the exact full conditionals for the switching and diffusion components.
Additionally, we sample the posterior parameters of the system by the use of conjugate updates.
In future work, an exciting direction is to extend the method to non-linear settings, where, e.g., samples from our method could serve as a proposal distribution in a particle smoother setting \cite{klaas2006fast,mider2021}.
Further, to allow for more expressive observation likelihoods, the latent hybrid process $\left\lbrace Z(t), Y(t)\right\rbrace$ could be combinded with recent neural network approaches to continuous-time processes \cite{li2020scalable}.

 Acknowledgements should only appear in the accepted version.
\section*{Acknowledgements}
We thank Tim Prangemeier for providing the gene switching data and helpful discussions and the anonymous reviewers for their useful comments and suggestions. This work has been funded by the European Research Council (ERC) within the CONSYN project, grant agreement number 773196, and by the German Research Foundation (DFG) as part of the project B4 within the Collaborative Research Center (CRC) 1053 – MAKI.


\bibliography{main}
\bibliographystyle{icml2021}


\appendix
\date{}
\newpage
\onecolumn
\title{Markov Chain Monte Carlo for Continuous-Time\\ Switching Dynamical Systems\\\bigskip\large{--- Supplementary Material ---}}
\maketitlenew
\section{Sampling the Conditional Diffusion Process}
\subsection{Derivation of the backward continuous-discrete Kalman filter}
\label{sec:app_backward_kalman}
The backward distribution $p(x_N\mid y, t)$ between the $N$th and $N-1$th observation is given by the \acf{kbe}, which reads
\begin{equation}
    \partial_t p(x_N\mid y, t) = -\mathcal{A}^\dagger p(x_N\mid y, t).
\end{equation}
Consider the linear dynamical case, here, the adjoint operator is given by
\begin{equation}
    \mathcal{A}^\dagger(\cdot)=\left(\nabla_y (\cdot)\right)^\top (A(t) y+b(t))+\frac{1}{2} \tr\left(D \nabla_y \nabla_y^\top(\cdot)\right).
\end{equation}
Hence, 
\begin{equation}
     \partial_t p(x_N\mid y, t)=-\left(\nabla_y p(x_N\mid y, t)\right)^\top (A(t) y+b(t))-\frac{1}{2} \tr\left(D \nabla_y \nabla_y^\top p(x_N\mid y, t)\right),
\end{equation}
where we assume the end-point condition
\begin{equation}
    p(x_N\mid y,T)=\NDis(x_N \mid F y,\Sigma).
\end{equation}
We write the \ac{kbe} component-wise as
\begin{equation}
     \partial_t p(x_N\mid y, t) = - \sum_{k,l} \partial_{y_k} p(x_N\mid y, t) A_{kl}(t) y_l - \sum_{k} \partial_{y_k} p(x_N\mid y, t) b_k(t)- \frac{1}{2} \sum_{k,l} \partial_{y_k}\partial_{y_l} p(x_N\mid y, t) D_{kl},
\end{equation}
We use the multivariate Fourier transform
\begin{equation}
    \hat{f}(u)=\mathcal{F}\{f(y)\}=\int f(y) e^{-i  u^\top y}\, \mathrm d y.
\end{equation}
Note the following rules 
\begin{equation}
\begin{aligned}
        &\mathcal{F} \{ f(Ly) \}=\frac{1}{\vert F \vert} \hat{f}(F^{-\top}u)\\
    &\mathcal{F} \{ \nabla_y f(y) \}=i u \hat{f}(u)\\
    &\mathcal{F} \{ y \nabla_y f(y) \}= i \nabla_u \hat{f}(u).
\end{aligned}
\end{equation}
Therefore, we find $\partial_t \hat{p}(x_N\mid u, t)$ as
\begin{equation}
\begin{aligned}
         \partial_t \hat{p}(x_N\mid u, t) &= -  \sum_{k,l} i \partial_{u_l} \mathcal{F} \{\partial_{y_k} p(x_N\mid y, t) A_{kl}(t) \}  -  \sum_{k} i u_k \hat{p}(x_N\mid u, t) b_k(t)- \frac{1}{2} \sum_{k,l} u_k u_l \hat{p}(x_N\mid u, t) D_{kl}\\
         &=-  \sum_{k,l} i \partial_{u_l} \{i u_k \hat{p}(x_N\mid u, t) A_{kl}(t) \}  -  \sum_{k} i u_k \hat{p}(x_N\mid u, t) b_k(t)- \frac{1}{2} \sum_{k,l} u_k u_l \hat{p}(x_N\mid u, t) D_{kl}\\
         &\,{}\begin{split}
             =\sum_{k,l} \partial_{u_l} u_k \hat{p}(x_N\mid u, t) A_{kl}(t)+ \sum_{k,l}  u_k \partial_{u_l} \hat{p}(x_N\mid u, t) A_{kl}(t) -  \sum_{k} i u_k \hat{p}(x_N\mid u, t) b_k(t)\\
             - \frac{1}{2} \sum_{k,l} i u_k i u_l \hat{p}(x_N\mid u, t) D_{kl}
         \end{split}\\
              &\,{}\begin{split}
             =\sum_{k,l} \1(k=l)\hat{p}(x_N\mid u, t) A_{kl}(t)+ \sum_{k,l}  u_k \partial_{u_l} \hat{p}(x_N\mid u, t) A_{kl}(t) -  \sum_{k} i u_k \hat{p}(x_N\mid u, t) b_k(t)\\
             + \frac{1}{2} \sum_{k,l} u_k u_l \hat{p}(x_N\mid u, t) D_{kl}.
         \end{split}\\
\end{aligned}
\end{equation}
This can be written in vector form as
\begin{equation}
    \partial_t \hat{p}(x_N\mid u, t)=\tr(A(t)) \hat{p}(x_N\mid u, t) + \left(\nabla_u \hat{p}(x_N\mid u, t) \right)^\top A^\top(t) u - i u^\top b(t) \hat{p}(x_N\mid u, t)  +\frac{1}{2} u^\top D u \hat{p}(x_N\mid u, t).
\end{equation}
We define the characteristic curve as 
\begin{equation}
    \frac{\mathrm{d}}{\mathrm d t} u(t)=-A^\top(t) u(t),
\end{equation}
with end-point boundary $u(T)=u_T$. 
The formal solution is
\begin{equation}
   u(t)=\Phi^\top(t,T) u_T.
\end{equation}
Hence, we have the following properties
\begin{equation}
\begin{aligned}
      u_T=\Phi^{-\top}(t,T) u(t)\\
     \frac{\mathrm{d}}{\mathrm d t} \Phi(t,T)=-\Phi(t,T) A(t).
\end{aligned}
\end{equation}
We build the total derivative of $\hat{p}(x_N\mid u, t)$ at $u=u(t)$ as
\begin{equation}
    \frac{\mathrm{d}}{\mathrm d t} \hat{p}(x_N\mid u(t), t)=\left( \nabla_u \hat{p}(x_N\mid u(t), t)\right)^\top  \frac{\mathrm{d}}{\mathrm d t} u(t)+\partial_t \hat{p}(x_N\mid u(t), t).
\end{equation}
Plugging in the Fourier transformed backward Fokker-Planck quantity $\partial_t \hat{p}(x_N\mid u(t), t)$ and the characteristic ODE $\frac{\mathrm{d}}{\mathrm d t} u(t)$, we find
\begin{equation}
    \frac{\mathrm{d}}{\mathrm d t} \hat{p}(x_N\mid u(t), t)=\left(\tr(A(t))  - i u^\top(t) b(t)  +\frac{1}{2} u^\top(t) D u(t)\right) \hat{p}(x_N\mid u(t), t).
\end{equation}
This yields the solution 
\begin{equation}
    \hat{p}(x_N\mid u(t), t)=\exp \left(-\int_{t}^T \tr(A(s))\, \mathrm d s  - i \left(-\int_{t}^T u^\top(s) b(s)\, \mathrm d s\right)  -\frac{1}{2} \int_{t}^T u^\top(s) D u(s)\, \mathrm d s\right)\hat{p}(x_N\mid u(T), T).
\end{equation}
For the end-point condition in Fourier space $\hat{p}(x_N\mid u(T), T)$, we compute
\begin{equation}
    \begin{aligned}
          &\hat{p}(x_N\mid u(T), T)=\hat{p}(x_N\mid u_T, T),
    \end{aligned}
\end{equation}
where we compute the Fourier transform as 
\begin{equation}
    \begin{aligned}
    \mathcal{F}\{p(x_N\mid y, T)\} =\mathcal{F}\{\NDis(x_N\mid F y,\Sigma)\}&=\mathcal{F}\{\NDis(F y\mid x_N,\Sigma)\}\\
    &=\frac{1}{\vert F \vert}\exp\left(-i \left(F^{-\top}u\right)^\top x_N -\frac{1}{2} \left(F^{-\top}u\right)^\top \Sigma F^{-\top}u\right),
    \end{aligned}
\end{equation}
where we assume $F$ is quadratic and invertible.
Hence we have
\begin{equation}
    \hat{p}(x_N\mid u(T), T)=\hat{p}(x_N\mid u_T, T)=\frac{1}{\vert F \vert}\exp\left(-i \left(F^{-\top}u_T\right)^\top x_N -\frac{1}{2} \left(F^{-\top}u_T\right)^\top \Sigma F^{-\top}u_T\right).
\end{equation}
Using the end-point condition for $ \hat{p}(x_N\mid u(t), t)$, we have
\begin{equation}
\begin{aligned}
        \hat{p}(x_N\mid u(t), t)=\frac{1}{\vert F \vert}\exp \left\{-\int_{t}^T \tr(A(s))\, \mathrm d s  - i \left(\left(F^{-\top}u_T\right)^\top x_N-\int_{t}^T u^\top(s) b(s)\, \mathrm d s\right)\right.  \\
        \left.-\frac{1}{2} \left(\left(F^{-\top}u_T\right)^\top \Sigma F^{-\top}u_T+\int_{t}^T u^\top(s) D u(s)\, \mathrm d s \right)\right\}.
\end{aligned}
\end{equation}
Using the formal solution $u(t)=\Phi^\top(t,T) u_T$ we have
\begin{equation}
\begin{aligned}
        \hat{p}(x_N\mid u(t), t)&=\frac{1}{\vert F \vert}\exp \left\{-\int_{t}^T \tr(A(s))\, \mathrm d s  - i \left(\left(F^{-\top}u_T\right)^\top x_N-u_T^\top \int_{t}^T \Phi(s,T)  b(s)\, \mathrm d s\right)\right.  \\
       &{} \left.-\frac{1}{2} \left(\left(F^{-\top}u_T\right)^\top \Sigma F^{-\top}u_T+u_T^\top \int_{t}^T \Phi(s,T)  D \Phi^\top(s,T)\, \mathrm d s \, u_T \right)\right\}\\
       &=\frac{1}{\vert F \vert}\exp \left\{-\int_{t}^T \tr(A(s))\, \mathrm d s  - i u_T^\top F^{-1}\left(x_N - F \int_{t}^T \Phi(s,T) b(s)\, \mathrm d s\right)\right.  \\
       &{} \left.-\frac{1}{2} u_T^\top F^{-1} \left( \Sigma + F \int_{t}^T \Phi(s,T)  D \Phi^\top(s,T)\, \mathrm d s \, F^\top \right)F^{-\top}u_T\right\}.
\end{aligned}
\end{equation}
Using $u_T=\Phi^{-\top}(t,T) u(t) \Longleftrightarrow u_T^\top =u^\top(t)\Phi^{-1}(t,T)$, we find
\begin{equation}
\begin{aligned}
          \hat{p}(x_N\mid u(t), t)&=\frac{1}{\vert F \vert}\exp \left\{-\int_{t}^T \tr(A(s))\, \mathrm d s  - i u^\top(t)\Phi^{-1}(t,T)  F^{-1}\left(x_N - F \int_{t}^T \Phi(s,T)  b(s)\, \mathrm d s\right)\right.  \\
       &{} \left.-\frac{1}{2} u^\top(t)\Phi^{-1}(t,T) F^{-1} \left( \Sigma + F \int_{t}^T \Phi(s,T)  D \Phi^\top(s,T)\, \mathrm d s \, F^\top \right)F^{-\top} \Phi^{-\top}(t,T)  u(t)\right\}\\
       &=\frac{1}{\vert F \vert}\exp \left\{-\int_{t}^T \tr(A(s))\, \mathrm d s  - i \left(\left( F \Phi(t,T)\right)^{-\top} u(t) \right)^\top \left(x_N - F \int_{t}^T \Phi(s,T)  b(s)\, \mathrm d s\right)\right.  \\
       &{} \left.-\frac{1}{2} \left(\left( F \Phi(t,T)\right)^{-\top} u(t)  \right)^\top \left( \Sigma + F \int_{t}^T \Phi(s,T)  D \Phi^\top(s,T)\, \mathrm d s \, F^\top \right)\left( F \Phi(t,T)\right)^{-\top} u(t) \right\}.
\end{aligned}
\end{equation}
Utilizing the inverse Fourier transform yields
\begin{equation}
\begin{aligned}
        p(x_N\mid y, t)&=\frac{\vert \Phi(t,T)\vert}{\exp \left(\int_{t}^T \tr(A(s))\, \mathrm d s\right)}
        \begin{multlined}[t]\NDis\left( F \Phi(t,T) y\,\middle|\, x_N - F \int_{t}^T \Phi(s,T)  b(s)\, \mathrm d s,\right.\\
        \qquad\qquad\qquad\qquad\Sigma + F \int_{t}^T \Phi(s,T)  D \Phi^\top(s,T)\, \mathrm d s\, F^\top \Bigg)\end{multlined}\\
        &=\frac{\vert \Phi(t,T)\vert}{\exp \left(\int_{t}^T \tr(A(s))\, \mathrm d s\right)}\begin{multlined}[t]\NDis\left( x_N \,\middle|\,  F \Phi(t,T) y  + F \int_{t}^T \Phi(s,T)  b(s)\, \mathrm d s,\right.\\
        \qquad\qquad\qquad\qquad\Sigma + F \int_{t}^T \Phi(s,T)  D \Phi^\top(s,T)\, \mathrm d s\, F^\top \Bigg).\end{multlined}
\end{aligned}
\end{equation}
Jacobi's formula yields
\begin{equation}
\begin{aligned}
    \frac{\mathrm d}{\mathrm d t} \vert \Phi(t,T) \vert &=\vert \Phi(t,T) \vert \tr\left(\Phi^{-1}(t,T) \frac{\mathrm d}{\mathrm d t}  \Phi(t,T) \right)\\
    &=\vert \Phi(t,T) \vert \tr\left(\Phi^{-1}(t,T)\left(-\Phi(t,T) A(t)\right)\right)\\
    &=\vert \Phi(t,T) \vert \tr\left(-A(t)\right).
\end{aligned}
\end{equation}
Therefore, we have for the solution 
\begin{equation}
     \vert \Phi(t,T) \vert=\exp\left(-\int_t^T \tr\left(-A(s)\right) \, \mathrm d s\right) \Phi(T,T)
\end{equation}
and $\Phi(T,T)=\mathbb I$.
Consequently,
\begin{equation}
     \vert \Phi(t,T) \vert=\exp\left(\int_t^T \tr\left(A(s)\right) \, \mathrm d s\right),
\end{equation}
and
\begin{equation}
\begin{aligned}
        p(x_N\mid y, t)&=\NDis\left( x_N \,\middle|\, F \Phi(t,T) y  + F \int_{t}^T \Phi(s,T)  b(s)\, \mathrm d s, \Sigma + F \int_{t}^T \Phi(s,T)  D \Phi^\top(s,T)\, \mathrm d s\, F^\top \right).
\end{aligned}
\end{equation}
Let $F(t)=F \Phi(t,T)$; computing the time derivative, we have
\begin{equation}
\begin{aligned}
    \frac{\mathrm d }{\mathrm d t} F(t)&=F  \frac{\mathrm d }{\mathrm d t} \Phi(t,T)\\
    &=F\left( -\Phi(t,T) A(t) \right )\\
    \Longleftrightarrow  \frac{\mathrm d }{\mathrm d t} F(t)&=-F(t) A(t),
\end{aligned}
\end{equation}
with end point condition $F(T)=F$.
Let now $m(t)=F \int_{t}^T \Phi(s,T)  b(s)\, \mathrm d s= \int_{t}^T F(s)  b(s)\, \mathrm d s$.
By differentiation (utilizing Leibniz' integral rule) we find
\begin{equation}
    \frac{\mathrm d }{\mathrm d t} m(t)=-F(t) b(t),
\end{equation}
with boundary condition $m(T)=0$.
Let further $\Sigma(t)=\Sigma + F \int_{t}^T \Phi(s,T)  D \Phi^\top(s,T)\, \mathrm d s\, F^\top = \Sigma + \int_{t}^T F(s)  D F^\top(s) \, \mathrm d s$.
We find analogously
\begin{equation}
    \frac{\mathrm d }{\mathrm d t} \Sigma(t)=-F(t) D F^\top(t),
\end{equation}
with boundary condition $\Sigma(T)=\Sigma$.

Summarizing, we have
\begin{equation}
            p(x_N\mid y, t)=\NDis\left( x_N \mid F(t) y  + m(t) , \Sigma(t) \right),
\end{equation}
with
\begin{equation}
    \begin{aligned}
    &\frac{\mathrm d }{\mathrm d t} F(t)=-F(t) A(t) && \text{with} &&&  F(T)=F,\\
    &\frac{\mathrm d }{\mathrm d t} m(t)=-F(t) b(t) && \text{with} &&&  m(T)=0,\\
     & \frac{\mathrm d }{\mathrm d t} \Sigma(t)=-F(t) D F^\top(t) && \text{with} &&&  \Sigma(T)=\Sigma.
    \end{aligned}
\end{equation}
\subsubsection{The proof for non-invertible $F$.}
In the proof we assumed that $F$ be invertible.
However, the solution also holds for general matrices $F$, which can be shown by plugging in the solution $p(x_N\mid y, t)=\NDis\left( x_N \mid F(t) y  + m(t) , \Sigma(t) \right)$ in the backward Fokker-Planck equation
\begin{equation}
         \partial_t p(x_N\mid y, t) = - \sum_{k,l} \partial_{y_k} p(x_N\mid y, t) A_{kl}(t) y_l - \sum_{k} \partial_{y_k} p(x_N\mid y, t) b_k(t)- \frac{1}{2} \sum_{k,l} \partial_{y_k}\partial_{y_l} p(x_N\mid y, t) D_{kl}.
\end{equation}
This yields the \ac{pde}
\begin{equation}
\begin{aligned}
                &\partial_t \NDis\left( x_N \mid F(t) y  + m(t) , \Sigma(t) \right) = - \sum_{k,l} \partial_{y_k} \NDis\left( x_N \mid F(t) y  + m(t) , \Sigma(t) \right) A_{kl}(t) y_l \\
                &{}- \sum_{k} \partial_{y_k} \NDis\left( x_N \mid F(t) y  + m(t) , \Sigma(t) \right) b_k(t)
                - \frac{1}{2} \sum_{k,l} \partial_{y_k}\partial_{y_l} \NDis\left( x_N \mid F(t) y  + m(t) , \Sigma(t) \right) D_{kl}.  
\end{aligned}
\end{equation}
We compute the partial derivatives $\partial_t \NDis\left( x_N \mid F(t) y  + m(t) , \Sigma(t) \right)$, $\partial_{y_k} \NDis\left( x_N \mid F(t) y  + m(t) , \Sigma(t) \right)$ and $\partial_{y_k} \partial_{y_l} \NDis\left( x_N \mid F(t) y  + m(t) , \Sigma(t) \right)$.
First, note the following
\begin{equation}
    \partial_\theta \NDis(x\mid a,A)=\NDis(x\mid a,A) \left(-h^\top (\partial_\theta x) + h^\top (\partial_\theta a)-\frac{1}{2}\tr(A^{-1}\partial_\theta A)+\frac{1}{2}h^\top(\partial_\theta A)h\right),
\end{equation}
with $h=A^{-1}(x-a)$.
This yields 
\begin{align}
&\begin{aligned}
             \partial_t \NDis\left( x_N \mid F(t) y  + m(t) , \Sigma(t) \right)=\NDis\left( x_N \mid F(t) y  + m(t) , \Sigma(t) \right) \left[h^\top (\partial_t F(t)y+\partial_t m(t))\right.\\
             \left.-\frac{1}{2}\tr\left(\Sigma^{-1}(t)\partial_t\Sigma(t)\right)+\frac{1}{2}h^\top \partial_t \Sigma h\right],
\end{aligned}\\
&\begin{aligned}
\partial_{y_k} \NDis\left( x_N \mid F(t) y  + m(t) , \Sigma(t) \right)=\NDis\left( x_N \mid F(t) y  + m(t) , \Sigma(t) \right) \left[h^\top F_{\cdot k}(t)\right],
\end{aligned}\\
&\begin{aligned}
\partial_{y_k} \partial_{y_l} \NDis\left( x_N \mid F(t) y  + m(t) , \Sigma(t) \right)&=\NDis\left( x_N \mid F(t) y  + m(t) , \Sigma(t) \right)  \left[h^\top F_{\cdot k}(t)\right]\left[h^\top F_{\cdot l}(t)\right]\\
&{}+\NDis\left( x_N \mid F(t) y  + m(t) , \Sigma(t) \right) \partial_{y_l}\left\{h^\top F_{\cdot k}(t)\right\}\\
&=\NDis\left( x_N \mid F(t) y  + m(t) , \Sigma(t) \right) \left[ h^\top F_{\cdot k}(t)h^\top F_{\cdot l}(t)- L^\top_{\cdot l}(t)\Sigma^{-1}(t)F_{\cdot k}(t)\right],
\end{aligned}
\end{align}
with $h=\Sigma^{-1}(t)(x_N-F(t)y-m(t))$
Inserting these equations into the \ac{kbe} yields
\begin{equation}
\begin{aligned}
       &h^\top (\partial_t F(t)y+\partial_t m(t))-\frac{1}{2}\tr\left(\Sigma^{-1}(t)\partial_t\Sigma(t)\right)+\frac{1}{2}h^\top \partial_t \Sigma h   \\
       &=-\sum_k h^\top F_{\cdot k,l}(t)A_{kl}(t) y_l
       -\sum_k h^\top F_{\cdot k}(t) b_k(t) 
       -\frac{1}{2}\sum_{k,l}\left(h^\top F_{\cdot k}(t) h^\top F_{\cdot l}(t)- F^\top_{\cdot l}(t)\Sigma^{-1}(t)F_{\cdot k}(t)\right)D_{kl}
\end{aligned}
\end{equation}
By utilizing vector notation, we have
\begin{equation}
\begin{aligned}
       &h^\top \left[\partial_t F(t)y+\partial_t m(t)\right]-\frac{1}{2}\tr\left\{\left(\Sigma^{-1}(t)-h h^\top\right) \partial_t\Sigma(t)\right\} \\
       &=h^\top \left[-F(t)A(t)y-F(t)b(t)\right]-\frac{1}{2}\tr\left\{\left(\Sigma^{-1}(t)-h h^\top \right)\left[-F^\top(t)D F(t)\right]\right\}
\end{aligned}
\end{equation}
By comparing coefficients. we find
\begin{equation}
    \begin{aligned}
    &\frac{\mathrm d }{\mathrm d t} F(t)=-F(t) A(t) && \text{with} &&&  F(T)=F\\
    &\frac{\mathrm d }{\mathrm d t} m(t)=-F(t) b(t) && \text{with} &&&  m(T)=0\\
     & \frac{\mathrm d }{\mathrm d t} \Sigma(t)=-F(t) D F^\top(t) && \text{with} &&&  \Sigma(T)=\Sigma,
    \end{aligned}
\end{equation}
where we find the end-point conditions by comparing the end-point boundary as
\begin{equation}
    \NDis(x_N\mid F(T)y+m(T),\Sigma(T))=\NDis(x_N\mid Fy,\Sigma).
\end{equation}

\paragraph{Reset conditions}
Starting at the end-point $t=T$, we consider the last observation $X_{N-1}$ at time point $t_{N-1}$.
We have, due to the Markov property,
\begin{equation}
\begin{split}
    \beta(y, t_{N-1}) = p(x_N, x_{N-1}\mid y, t_{N-1}) &= p(x_N\mid y, t_{N-1}) p(x_{N-1}\mid x_N, y, t_{N-1})\\
    &= p(x_N\mid y, t_{N-1}) p(x_{N-1}\mid y, t_{N-1})\\
    &= \beta(y, t^+_{N-1}) p(x_{N-1}\mid y, t_{N-1}),
    \end{split}
\end{equation}
where $\beta(y, t^+_{N-1}) := \lim_{h\searrow0}\beta(y, t_{N-1} + h)$.
As we assume a Gaussian observation likelihood, we have, due to the Gaussian properties,
\begin{align}
    \beta(y, t_{N-1}) &= \NDis\left(x_N\mid F(t_{N-1}^+)y + m(t_{N-1}^+), \Sigma(t_{N-1}^+\right)\NDis(x_{N-1}\mid y , \Sigma_{x})\\
    &= \NDis\left(x_{N-1}, x_N\mid F(t_{N-1})y + m(t_{N-1}), \Sigma(t_{N-1})\right)
\end{align}
with reset parameters
\begin{align}\label{eq:reset_conditions_kalman}
    F(t_{N-1}) &= \begin{pmatrix}
 \mathbb I_n \\
 F(t_{N-1}^+)
\end{pmatrix} \in \R^{2n\times n},\\
m(t_{N-1}) &= \begin{pmatrix}
    0_n\\
    m(t_{N-1}^+)
\end{pmatrix}\in \R^{2n},\\
M(t_{N-1}) &= \begin{pmatrix}
\Sigma_{x} & 0 \\
0 & \Sigma(t_{N-1}^+)
\end{pmatrix}\in \R_{\mathrm{psd}}^{2n \times 2n},
\end{align}
where $0_n$ denotes the $n$-dimensional all-zeros vector and $\mathbb I_n$ is the $n$-dimensional identity matrix.

\subsubsection{Information filter parameterization}
The above backward filter has the property that its support increases upon every incorporated observation, which is computationally disadvantageous.  
Notice, however, that the contribution of the backward filter to the drift of the posterior \ac{sde} is fixed in size; for convenience, we re-state the \ac{sde}:
\begin{equation}
        \mathrm d Y(t) = \left(f(Y(t), t) + D(Z(t)) \nabla \log \beta(Y(t), t)\right)\mathrm dt + Q(Z(t))\mathrm d W(t),\nonumber
\end{equation}
where we notice that 
\begin{align}
    \nabla \log \beta(y, t) &= -\frac{1}{2} \nabla (x(t) - F(t)y(t) - m(t))^\top \Sigma^{-1}(t)(x(t) - F(t)y(t) - m(t))\\
    &= F(t)^\top \Sigma^{-1}(t)(x(t) - m(t)) - F(t)^\top \Sigma^{-1}(t) F(t)y(t)
\end{align}
where the gradient was taken with respect to $y$.
Defining
\begin{equation}\label{eq:conversion_kalman_information_filter}
\begin{split}
    \nu(t) &:= F(t)^\top \Sigma^{-1}(t)(x(t) - m(t)),\\
    M(t) &:= F(t)^\top \Sigma^{-1}(t) F(t),
    \end{split}
\end{equation}
one can compute straightforwardly the respective time derivatives, where we use the notation $\dot{f} = \frac{\mathrm d }{\mathrm d t}f$ for conciseness,
\begin{align}
\begin{split}
    \frac{\mathrm d}{\mathrm d t}M(t) &= \dot{F}(t)^\top \Sigma^{-1}(t) F(t) + F(t)^\top \dot{\Sigma}^{-1}(t) F(t) + F(t)^\top \Sigma^{-1}(t) \dot{F}(t)\\
    &= -A(t)^\top M(t) + F(t)^\top \dot{\Sigma}^{-1}(t)F(t) - M(t)A(t)\\
    &= -A(t)^\top M(t) - F(t)^\top \Sigma^{-1}(t) \dot{\Sigma}(t) \Sigma^{-1}(t)F(t) - M(t)A(t)\\
    &= -A(t)^\top M(t) + F(t)^\top \Sigma^{-1}(t) F(t) D F(t)^\top \Sigma^{-1}(t)F(t) - M(t)A(t)\\
    &= -A(t)^\top M(t) + M(t) D M(t) - M(t)A(t),
    \end{split}\\
    \begin{split}
        \frac{\mathrm d }{\mathrm d t } \nu (t)&=  \dot{F}(t)^\top \Sigma^{-1}(t)(x(t) - m(t)) + F(t)^\top \dot{\Sigma}^{-1}(t)(x(t) - m(t)) - F(t)^\top\Sigma^{-1}(t)\dot{m}(t)\\
        &= -A(t)^\top \nu(t) + M(t)D\nu(t) + M(t)b(t).
    \end{split}
\end{align}

The reset conditions for the information filter given in the main paper follow directly by comparing \cref{eq:reset_conditions_kalman} and \cref{eq:conversion_kalman_information_filter}.

\pagebreak
\subsection{Posterior \ac{ssde} Initial Condition}
\label{sec:app_posterior_ssde_init}
The posterior initial distribution $p(y_0\mid x_{[1,N]})$ is found as
\begin{equation}
\begin{split}
p(y_0\mid x_{[1, N]}, -) &\propto \beta(y_0, 0)p(y_0\mid \mu_0, \Sigma_0)\\
&\propto \exp\left\lbrace-c(0) - \frac{1}{2}y_0^\top I(0)y_0 + a(0)^\top y_0\right\rbrace\NDis\!\left(y_0\mid \mu_0, \Sigma_0\right)\\
&\propto \NDis\!\left(y_0\mid \bar{\mu}, \bar{\Sigma}\right)
\end{split}
\end{equation}
with
\begin{equation}
    \bar{\mu} = \bar{\Sigma} (\Sigma_0^{-1}\mu_0 + a(0)),\quad \bar{\Sigma} = \left(\Sigma_0^{-1} + I(0)\right)^{-1}.
\end{equation}

\section{Derivation of the Wonham-type filter}
\label{sec:app_kushner-stratonovich}
In the following, we derive the filtering density $p_f(z, t)$.
We follow the treatment in \cite{del2017stochastic}, but see also, e.g., \cite{vanhandelFilteringStabilityRobustness2007}.
For convenience, we re-iterate the time evolution of the hybrid system as
\begin{equation}
\begin{split}
\frac{\mathrm d}{\mathrm dt}p(z, t) &= \sum_{z'\in \mathcal Z}\Lambda(z', z, t)p(z', t)\quad\forall z' \in \mathcal Z,\\
\mathrm dY(t) &= f(Y(t), Z(t))\mathrm dt + Q(Z(t))\mathrm dW(t),\\
\end{split}
\end{equation}
where the dispersion $Q(z)Q(z)^\top = D(z)$.
We are interested in the conditional path measure 
\begin{equation}
\begin{split}
\mathsf P\left(Z_{[0,t]}\in \mathrm dz_{[0,t]} \mid y_{[0,t]}\right)&\propto \mathsf P\left((Z_{[0,t]}, Y_{[0,t]}) \in \mathrm d (z_{[0,t]}, y_{[0,t]})\right)\\
&= \mathsf P\left(Y_{[0,t]}\in \mathrm dy_{[0,t]}\mid z_{[0,t]}\right)\mathsf P\left(Z_{[0,t]}\in \mathrm dz_{[0,t]}\right)\\
&= G(z_{[0,t]}, y_{[0,t]})\mathsf P(W_{[0,t]}\in \mathrm dy_{[0,t]})\mathsf P\left(Z_{[0,t]}\in \mathrm dz_{[0,t]}\right)
\end{split}
\end{equation}
where the last equality is due to Girsanov's theorem \cite{oksendal2003stochastic} with the Radon-Nikodym derivative
\begin{equation}
    G(z_{[0,t]}, y_{[0, t]}) := \frac{\mathrm d \mathsf P_{Z}}{\mathrm d \mathsf P_{W}} = \exp\left\lbrace F(t)\right\rbrace,
\end{equation}
where we used the subscripts to indicate the conditional posterior and the Brownian motion measures and defined the shorthand
\begin{multline}
    F(t) = \int_0^t \mathrm d F(t) := \int_0^t f^\top(z(s), y(s))D(z(s))^{-1}\mathrm dy(s)
    - \frac{1}{2}\int_0^t f^\top(z(s), y(s))D(z(s))^{-1}f(z(s), y(s))\mathrm ds,
\end{multline}
the posterior measure can be expressed via the Kallianpur-Striebel formula,
\begin{equation}
    \mathsf P\left(Z_{[0,t]}\in \mathrm dz_{[0,t]} \mid y_{[0,t]}\right) = \frac{G(z_{[0,t]}, y_{[0, t]})\mathsf P(Z_{[0,t]}\in \mathrm dz_{[0,t]})}{\int G(z'_{[0,t]}, y_{[0, t]})\mathsf P(Z_{[0,t]}\in \mathrm dz'_{[0,t]})}.
    \label{eq:kallianpur-striebel}
\end{equation}
Our quantity of interest follows as an expectation,
\begin{equation}
    p_f(z, t) = \E\left[\mathbbm 1(Z(t) = z)\right] = \int \mathbbm 1(Z(t)=z)\mathsf P\left(Z_{[0,t]} \in \mathrm dz_{[0,t]}\mid y_{[0,t]}\right),
\end{equation}
where we use the subscript $f$ to indicate the \emph{filtering} distribution.
Using \cref{eq:kallianpur-striebel}, we can restate this in terms of the prior measure, 
\begin{equation}\label{eq:filtering_distribution}
    p_f(z, t) = \frac{\int \mathbbm 1(Z(t)=z) G(z_{[0,t]}, y_{[0, t]})\mathsf P(Z_{[0,t]}\in \mathrm dz_{[0,t]})}{\int G(z'_{[0,t]}, y_{[0, t]})\mathsf P(Z_{[0,t]}\in \mathrm dz'_{[0,t]})} = \frac{\E\left[\mathbbm 1(Z(t) = z)G(z_{[0,t]}, y_{[0,t]})\right]}{\E\left[G(z_{[0,t]}, y_{[0,t]})\right]}.
\end{equation}
Here and in the following, the expectation operator $\E[\cdot]$ refers to the expectation over the prior measure $P(Z_{[0,t]}\in \mathrm dz_{[0,t]})$.
Using the It\^{o} calculus chain rule, we compute 
\begin{equation}
\begin{split}
    \mathrm d G(z_{[0,t]}, y_{[0,t]}) &= \exp\lbrace F(t)\rbrace \mathrm d F(t) + \frac{1}{2}\exp\lbrace F(t)\rbrace \mathrm d F(t)\mathrm d F(t)\\
    &=G(z_{[0,t]}, y_{[0,t]}) \mathrm d F(t) + \frac{1}{2}G(z_{[0,t]}, y_{[0,t]})\mathrm d F(t)\mathrm d F(t)\\
    &= G(z_{[0,t]}, y_{[0,t]})f(z(t), y(t))^\top D(z(t))^{-1}\mathrm dy(t).
    \end{split}
\end{equation}
and find for the unnormalized quantity 
\begin{equation}
    \tilde{p}_f(z, t) := \E\left[\mathbbm 1(Z(t) = z)G(z_{[0,t]}, y_{[0,t]})\right]
\end{equation}
the expression
\begin{equation}
\begin{split}
    \mathrm d \tilde{p}_f(z, t) &= \E\left[\mathrm d\!\left(\mathbbm 1(Z(t) = z) G(z_{[0,t]}, y_{[0, t]})\right)\right]\\
    &= \E\left[\mathrm d \mathbbm 1(Z(t) = z)G(z_{[0,t]}, y_{[0,t]}) + \mathbbm 1(Z(t) = z)\mathrm d G(z_{[0,t]}, y_{[0,t]}) + \mathrm d \mathbbm 1(Z(t) = z)\mathrm d G(z_{[0,t]}, y_{[0,t]})\right].
\end{split}
\end{equation}
Now, noticing that
\begin{equation}
    \mathrm d\mathbbm 1(Z(t)=z) = \sum_{z'\in \mathcal Z}\Lambda(z, z', t)\mathbbm 1(Z(t) = z)\mathrm dt
\end{equation}
and accordingly, acknowledging that $\mathrm d t \cdot \mathrm dy(t) = 0$,
\begin{equation}\label{eq:zakai}
\begin{split}
    \mathrm d \tilde{p}_f(z, t) &= \E\left[\mathrm d \mathbbm 1(Z(t) = z)G(z_{[0,t]}, y_{[0,t]}) + \mathbbm 1(Z(t) = z)\mathrm d G(z_{[0,t]}, y_{[0,t]})\right]\\
    &= \sum_{z'\in\mathcal Z}\Lambda(z', z, t)\tilde{p}_f(z', t)\mathrm d t + \E\left[\mathbbm 1(Z(t) =z)f^\top(y(t), z(t))D^{-1}(z(t))\right]\mathrm d y(t)\\
    &= \sum_{z'\in\mathcal Z}\Lambda(z', z, t)\tilde{p}_f(z', t)\mathrm d t + \tilde{p}_f(z, t)f^\top(y(t), z(t))D^{-1}(z(t))\mathrm d y(t).
\end{split}
\end{equation}
\Cref{eq:zakai} is called the Zakai equation in the literature.
To derive the dynamics of the desired respective normalized quantity \cref{eq:kallianpur-striebel}, further consider its denominator and notice that
\begin{equation}\label{eq:dlogE}
     \mathrm d\!\log\E\!\left[G\right]= \frac{\mathrm d\! \E\!\left[G\right]}{\E\!\left[G\right]} - \frac{1}{2}\frac{\tr\left\lbrace\mathrm d\! \E\!\left[G\right]\mathrm d \!\E\!\left[G\right]^\top\right\rbrace}{\E\!\left[G\right]^2},
\end{equation}
where we suppressed the arguments for conciseness. 
The quantity $\mathrm d\E\!\left[G\right]$ is precisely given by the Zakai equation \labelcref{eq:zakai} upon replacing the indicator by a constant, $\mathbbm 1(Z(t)=z) \rightarrow 1$, 
\begin{equation}
    \mathrm d\!\E\!\left[G\right] = \E\left[Gf^\top D^{-1}\right]\mathrm d y(t).
\end{equation}
Inserting this into \cref{eq:dlogE}, as $\mathrm dy(t) \mathrm dy(t)^\top = D(z(t))\mathrm dt$, one finds
\begin{equation}
\begin{split}
     \mathrm d\!\log\E\!\left[G\right]&= \frac{\E\!\left[G f^\top D^{-1}\right]}{\E\!\left[G\right]} \mathrm dy(z)- \frac{1}{2}\frac{\E\!\left[G f^\top D^{-1} \right]D^{-1}\E\!\left[D^{-1}f \right]}{\E\!\left[G\right]^2}\mathrm d t
\end{split}
\end{equation}
where the terms on the right hand side are of the exact same form as \cref{eq:filtering_distribution}.
Consequently,
\begin{equation}
    \E\!\left[G\right] = \exp\bigg\lbrace\int_0^t\underbrace{\frac{\E\!\left[G f^\top D^{-1} \right]}{\E\!\left[G\right]}}_{=:\varpi^\top} \mathrm d y(s)- \frac{1}{2}\int_0^t\underbrace{\frac{\E\!\left[G f^\top D^{-1}\right]D^{-1}\E\!\left[D^{-1}fG \right]}{\E\!\left[G\right]^2}}_{\varpi^\top D^{-1}\varpi}\mathrm d s\bigg\rbrace,
\end{equation}
where, for clarity, we restate with arguments:
\begin{equation}
    \varpi(t) = \frac{\E\!\left[G(z_{[0,t]}, y_{[0,t]}) f(y(t), z(t))^\top D(z(t))^{-1} \right]}{\E\!\left[G(z_{[0,t]}, y_{[0,t]})\right]}.
\end{equation}
Inserting this into the original \cref{eq:filtering_distribution}, we arrive at
\begin{multline}
    p_f(z, t) = \E\left[\mathbbm 1(Z(t) = z)\exp\left\lbrace\int_0^t \left(f^\top(z(s), y(s))D(z(s))^{-1} - \varpi\right)\left(\mathrm dy(s) - \varpi(s)\right)\right.\right.\\
    \left.\left.- \frac{1}{2}\int_0^t \left(f^\top(z(s), y(s)) - \varpi(s)\right)D(z(s))^{-1}\left(f(z(s), y(s)) - \varpi(s)\right)\mathrm ds\right\rbrace\right].
\end{multline}
Repeating with this quantity the same derivation steps as for the Zakai equation \labelcref{eq:zakai} straightforwardly yields the Kushner-Stratonovich equation
\begin{equation}
    \mathrm d p_f(z, t) = \sum_{z'\in \mathcal Z}\Lambda(z', z, t)p_f(z, t) \mathrm d t + p_f(z, t)\left(f(y(t), z(t) - \bar{f}(y(t))\right)^\top D(z(t))^{-1}\left(\mathrm d y(t) - \bar{f}(y(t))\right),
\end{equation}
where $\bar{f}(y(t)):= \sum_{z\in\mathcal Z}f(y(t), z(t))p_f(z, t)$.

\section{Bayesian Parameter estimation}
\label{sec:app_parameter_updates}
We go through the derivations in the same order as presented in the main paper.

\paragraph{Initial Conditions}
The Dirichlet prior in the initial \ac{mjp} state distribution $\pi_{z_0}$,
\begin{equation}
    p(\pi_{z_0}) = \DirDis(\pi_{z_0}\mid \alpha_{z_0}),
\end{equation}
with $\alpha_{z_0} \in \R_{>0}^{\vert\mathcal Z\vert}$, results in the usual conjugate update, 
\begin{equation}
\begin{split}
p(\pi_{z_0}\mid z_{[0,T]}, -) &\propto \CatDis(z(0)\mid \pi_{z_0})\DirDis(\pi_{z_0}\mid\alpha_{z_0})\\
&\propto \DirDis(\pi_{z_0}\mid \alpha_{z_0} + \delta_{z(0)}).
\end{split}
\end{equation}
The \ac{ssde} initial distribution parameters $\mu_0, \Sigma_0$ are specified via a \ac{niw} prior, 
\begin{equation}
    \NIWDis(\mu_0, \Sigma_0\mid \eta, \lambda, \Psi, \kappa) = \NDis\left(\mu_0\,\middle|\, \eta, \frac{\Sigma_0}{\lambda}\right)\IWDis(\Sigma_0\mid \Psi, \kappa).
\end{equation}
Recall that the \ac{ssde} initial value $y(0) = y_0$ is Gaussian distributed,
\begin{equation}
\begin{split}
    p(y_0) &= \NDis\!\left(y_0\mid \mu_0, \Sigma_0\right).
    \end{split}
\end{equation}
Accordingly, the Bayesian posterior
\begin{equation}
\begin{split}
    p(\mu_0, \Sigma_0\mid y_0, x_{[1, N]}, -) &\propto p(y_0\mid x_{[1,N]}, \mu_0, \Sigma_0)p(\mu_0, \Sigma_0)\\
    &\propto p(y_0\mid \mu_0, \Sigma_0)p(\mu_0, \Sigma_0)\\
    &\propto \NIWDis\!\left(\mu_0, \Sigma_0\,\middle|\,\tilde{\eta}, \tilde{\lambda}, \tilde{\Psi}, \tilde{\kappa}\right)
\end{split}
\end{equation}
with the updates as in the main paper.

\paragraph{MJP rate parameters}
Utilizing the transition counts and the cumulative sojourn times as defined in the main paper,
\begin{equation}
\begin{split}
     N_{zz'}&=\sum_{k=0}^{K-1} \1(z_k=z,z_{k+1}=z'),\\
     T_z&=\sum_{k=0}^K \1(z_k=z)\tau_k,
\end{split}
\end{equation}
we can explicate the path likelihood $p(z_{[0,T]}\mid \Lambda_{zz'})$ and compute
\begin{equation}
\begin{aligned}
   p(\Lambda_{z z'} \mid z_{[0,T]}) &\propto p(z_{[0,T]}, -\mid \Lambda_{z z'}) p(\Lambda_{z z'}) \\
 &\propto \prod_{k=0}^{K-1} \left\{\Lambda_{z_k} e^{-\Lambda_{z_k} \tau_k} \right\}^{\1(z_k=z)} \left\{\frac{\Lambda_{z_k z_{k+1}}}{\Lambda_{z_k}}\right\}^{\1(z_k=z, z_{k+1}=z')} \{e^{-\Lambda_{z_K} \tau_K}\}^{\1(z_K=z)} p(\Lambda_{z z'})\\
 &\propto \left\{\Lambda_{z z'}\right\}^{\sum_{k=0}^{K-1} \1(z_k=z,z_{k+1}=z')} e^{-\Lambda_z \sum_{k=0}^K \1(z_k=z) \tau_k} p(\Lambda_{z z'})\\
 &\propto \left\{\Lambda_{z z'}\right\}^{\sum_{k=0}^{K-1} \1(z_k=z,z_{k+1}=z')} e^{-\Lambda_{z z'} \sum_{k=0}^K \1(z_k=z) \tau_k} p(\Lambda_{z z'})\\
 &\propto \GamDis(\Lambda_{z z'} \mid s +N_{zz'},r+T_z ).
\end{aligned}
\end{equation}


\paragraph{\ac{sde} Drift Parameters}
We impose a \acf{mn} prior over the drift parameters $\Gamma_m:=\left[A_m, b_m\right] $, 
\begin{equation}
\begin{split}
    p(\Gamma_m) &= \MNDis(\Gamma_m\mid M_m, D_m, K_m)\\
    \MNDis(\Gamma_m\mid M_m, D_m, K_m) &=  (2\pi)^{\frac{nm}{2}}\vert D_m \vert^{-\frac{n}{2}}\vert K_m \vert^{-\frac{m}{2}}\exp\left\lbrace-\frac{1}{2}\tr{\left((\Gamma_m - M_m)^\top D_m^{-1}(\Gamma_m - M_m)K_m^{-1}\right)}\right\rbrace.
    \end{split}
\end{equation}
Note that we atypically use the subscript $m$ for `mode' here to avoid visual confusion with the letter $z$.
To update the drift parameters $\Gamma_m$ for all modes $m\in \mathcal Z$, we need to compute
\begin{align}
    p(\Gamma_m\mid z_{[0,T]}, y_{[0,T]}, x_{[1,N]}, -) &\propto G(z_{[0,T]}, y_{[0,T]}\mid \Gamma_m)p(\Gamma_m).
\end{align}
The `likelihood term' can be evaluated approximately by inserting the simulated paths.
Now note that for the mode $m$, only the subintervals of $z_{[0,T]}$ contribute in which $z(t) = m$.
More concretely, an \ac{mjp} realization $z_{[0,T]}$ can be specified via the \emph{jump times} $\left\lbrace j_k \right\rbrace_{k=1,...,K}$,
\begin{equation}
    j_{k+1} = \inf_t\left(t \geq j_k \mid z(t) \neq z(j_k)\right)
\end{equation}
and the \emph{state sequence} $\left\lbrace z_k\right\rbrace_{k=1,...,K},\, z_k\in \mathcal{Z}\, \forall k$:
\begin{equation}
    z(t) = z_{i},\, i = \sup_k(j_k < t).
\end{equation}
This allows us to write
\begin{align}
    p(\Gamma_m\mid z_{[0,T]}, y_{[0,T]}, x_{[1,N]}, -) &\propto G(z_{[0,T]}, y_{[0,T]}\mid \Gamma_m)p(\Gamma_m)\\
    &=\exp\left\lbrace\sum_{i}\int_{j_{m_i}}^{j_{m_i+1}} f^\top(z(s), y(s))D^{-1}(z(s))\mathrm dy(s)\right.\nonumber\\
    &\qquad\qquad\left.- \frac{1}{2}\int_{j_{m_i}}^{j_{m_i+1}} f^\top(z(s), y(s))D^{-1}(z(s))f(z(s), y(s))\mathrm ds\right\rbrace p(\Gamma_m),
\end{align}
where the summation goes over all $i$ for which $z(j_{m_i}) = m$.
Putting this sum over intervals aside for a moment for better readability, we find upon inserting a simulated \ac{sde}-path $y_{[0,T]}$ 
\begin{align}
    &\exp\left\lbrace\int_{t_0}^{t_1} f^\top(m, y(s))D^{-1}(m)\mathrm dy(s)- \frac{1}{2}\int_{t_0}^{t_1} f^\top(m, y(s))D^{-1}(m)f(m, y(s))\mathrm ds\right\rbrace \\
    &\approx \exp\left\lbrace\sum_{l=1}^L f^\top(m, y_l)D^{-1}(m) \Delta y_l- \frac{1}{2} \sum_{l=1}^L f^\top(m, y_l)D^{-1}(m)f(m, y_l)h\right\rbrace,
\end{align}
where $h$ is time simulation time-step, $s_l = s_{l-1} + h$, the interval boundaries $s_1 = t_0$, $s_L = t_1$, and $\Delta y_l := y(s_l + h) - y(s_l)$ the difference of two successive points of the trajectory.

Inserting the drift $f(m, y_l) = \Gamma_m\bar{y}_l$, where $\bar{y}_l = \left[y_l^\top, 1^\top_n \right]^\top$, and writing for conciseness $D^{-1}(m) = D^{-1}_m$ we find
\begin{align}
    &\exp\left\lbrace\sum_{l=1}^L f^\top(m, y_l)D^{-1}_m \Delta y_l- \frac{1}{2} \sum_{l=1}^L f^\top(m, y_l)D^{-1}_m f(m, y_l)h\right\rbrace\\
    &\quad =\exp\left\lbrace\sum_{l=1}^L \bar{y}^\top_l \Gamma_m^\top D^{-1}_m \Delta y_l- \frac{1}{2} \sum_{l=1}^L \bar{y}^\top_l \Gamma_m^\top D^{-1}_m\Gamma_m \bar{y}_l h\right\rbrace\\
    &\quad =\exp\left\lbrace\sum_{l=1}^L \sqrt{h}\bar{y}^\top_l \Gamma_m^\top D^{-1}_m \frac{\Delta y_l}{\sqrt{h}}- \frac{1}{2} \sum_{l=1}^L \sqrt{h}\bar{y}^\top_l \Gamma_m^\top D^{-1}_m\Gamma_m \bar{y}_l \sqrt{h}\right\rbrace\\
    &\quad =\exp\left\lbrace -\frac{1}{2}\sum_{l=1}^L\left(\frac{\Delta y_l}{\sqrt{h}} - \Gamma_m \bar{y}_l \sqrt{h}\right)^\top D^{-1}_m \left(\frac{\Delta y_l}{\sqrt{h}} - \Gamma_m \bar{y}_l \sqrt{h}\right) + \frac{1}{2}\sum_{l=1}^L \Delta y_l^\top D^{-1}_m \Delta y_l \frac{1}{h}\right\rbrace.
\end{align}
We may now omit the last term on the right hand side, as it is independent of $\Gamma_m$.
Note, however, that together with the measure of the Brownian motion in \cref{eq:conditional_Z_measure}, we have \cite{del2017stochastic}
\begin{equation}
    \mathsf P\!\left(W_{[0,T]}\in \mathrm dw_{[0,T]}\right) \exp\left\lbrace \frac{1}{2}\sum_{l=1}^L \Delta y_l^\top D^{-1}_m \Delta y_l \frac{1}{h}\right\rbrace = \vert 2\pi h D_m\vert^{-\frac{1}{2}}\prod_{l=1}^L \Delta y_l.
\end{equation}
Hence, the above is equivalent to approximating the Radon-Nikodym derivative $G(z_{[0,T]}, y_{[0,T]})$ via the product of $L$ Gaussian transition distributions.
Making use of the trace function and defining the joint observation vectors
\begin{align}
    \Delta Y &:= \left[\frac{\Delta y_{s_1}}{\sqrt{h}}, \cdots ,\frac{\Delta y_{s_L}}{\sqrt{h}}\right]\in \R^{n\times L},\\
    \bar{Y} &:= \left[\bar{y}_{s_1}\sqrt{h}, \cdots ,\bar{y}_{s_1}\sqrt{h}\right]\in \R^{n+1\times L},
\end{align}
we arrive at
\begin{align}
    &\exp\left\lbrace -\frac{1}{2}\sum_{l=1}^L\left(\frac{\Delta y_l}{\sqrt{h}} - \Gamma_m \bar{y}_l \sqrt{h}\right)^\top D^{-1}_m \left(\frac{\Delta y_l}{\sqrt{h}} - \Gamma_m \bar{y}_l \sqrt{h}\right)\right\rbrace\\
    &\quad = \exp\left\lbrace -\frac{1}{2}\tr{\left(\left(\Delta Y - \Gamma_m \bar{Y}\right)^\top D^{-1}_m\left(\Delta Y - \Gamma_m \bar{Y}\right)\mathbbm 1_{L\times L}\right)}\right\rbrace
\end{align}
and notice that this expression corresponds to an (un-normalized) \ac{mn} distribution 
Accordingly, still only considering a single jump interval $[j_{m_i}, j_{m_i + 1}]$, we can write 
\begin{align}
    p(\Gamma_m\mid z_{[0,T]}, y_{[0,T]}, x_{[1,N]}, -) &\propto \MNDis(\Delta Y \mid \Gamma_m \bar{Y}, D_m, \mathbbm 1_{L\times L}) p(\Gamma_m)\\
    &=\MNDis(\Delta Y \mid \Gamma_m \bar{Y}, D_m, \mathbbm 1_{L\times L})\MNDis(\Gamma_m \mid M_m, D_m, K_m)
\end{align}
It is known \cite{willskyNonparametricBayesianLearning2009} that the the Matrix-Normal distribution is a conjugate prior to the Matrix-Normal likelihood in the above form; consequently, the sought-after posterior is itself Matrix-Normal
\begin{equation}
    p(\Gamma_m, z_{[0,T]}, y_{[0,T]}, x_{[1,N]}, -) = \MNDis(\Gamma_m \mid \tilde{M}_m, D_m, \tilde{K}_m)
\end{equation}
with posterior hyperparameters
\begin{equation}
\begin{split}
    \tilde{K}_m&=\bar{Y}\bar{Y}^\top + K_m,\\
    \tilde{M}_m&=(\Delta Y\bar{Y}^\top + M_m K_m)\tilde{K}_m^{-1}.\\
 \end{split}
\end{equation}
Summation over all intervals with $z(j_{m_i}) = m$ is straightforward.
Importantly, the above derivation also holds for adaptive step-sizes, $s_l = s_{l-1} + h_{l-1}$.


\paragraph{Observation Covariance}
We define the prior
\begin{equation}
    p(\Sigma_{x}\mid x_{[1,N]}) = \IWDis(\Sigma_{x}\mid \Psi_{x}, \lambda_{x}).
\end{equation}
With the Gaussian observations
$$
x_i \sim \NDis(x_i\mid y_i, \Sigma_{\mathrm{obs}}),
$$
the standard result is
\begin{align}
p(\Sigma_{\mathrm{obs}}\mid x_{[1,N]}) &\propto p(x_{[1,N]}\mid \Sigma_{\mathrm{obs}})p(\Sigma_{\mathrm{obs}})\\
&\propto \IWDis(\Sigma_{\mathrm{obs}}\mid \tilde{\Psi}_{\mathrm{obs}}, \tilde{\lambda}_{\mathrm{obs}})
\end{align}
with the updated hyperparameters as in the main paper.

\section{Experimental Details}
\label{sec:app_experimental_details}
\subsection{Hyperparameter Settings}
We initialize all distribution hyperparameters, cf. \cref{sec:app_parameter_updates}, empirically.
To this end, we run k-means with the number of modes $\vert \mathcal Z \vert$ on the data and obtain empirical cluster means $\mu_z$ and covariances $\Sigma_z$.
In the following, we denote by $z(t_i)$, we mean the k-means cluster assignment of observation $i$ at time point $t_i$.
We go through the settings in order of appearance in the main paper.
\paragraph{Initial Conditions}
The \ac{mjp} initial Dirichlet hyperparameters $$\alpha_z=1 + \delta(z(t_1))$$.
The \ac{sde} initial \ac{niw} hyperparameters
\begin{equation*}
        \eta  = \frac{\sum_z \mu_z}{\vert Z\vert},\quad
        \lambda = 1,\quad
        \Psi = \frac{\sum_z \Sigma_z}{\vert Z\vert} * 0.1,\quad
        \kappa = n + 2.
\end{equation*}
Note that, as done, e.g., in \cite{willskyNonparametricBayesianLearning2009}, we use a heuristic downscaling of the empirical covariances as they contain contributions by the measurement noise, the process covariance as well as the drift.
Also, $\kappa=n+2$ is the smallest scaling parameter that makes the \ac{iw} distribution well defined.

\paragraph{\ac{mjp} rates}
We compute the number of total observed transitions in the k-means trajectory, $N_{\mathrm{trans}}$ and set
\begin{equation*}
    s = N_{\mathrm{trans}},\quad r = 1.
\end{equation*}

\paragraph{\ac{sde} Drift Parameters}
We compute 
\begin{equation}
\begin{split}
    \hat{A}_z &= \sum_{i=1}^N \mathbbm 1(z(t_i) = z)\frac{x_{i+1} - x_i}{x_{i+1} - t_i},\\
    \hat{b}_z &= -\hat{A}_z\mu_z,
    \end{split}
\end{equation}
where the latter is because the set point for a linear system is found via
\begin{equation}
    f(y) = Ay + b = A(y + A^{-1}b),
\end{equation}
hence $f(y) = 0$ if $y = -A^{-1}b$, and we demand 
\begin{equation}
    \mu_z = -A_z^{-1}b\,\Rightarrow b_z = -A_z\mu_z.
\end{equation}
With this, the \ac{mn} hyperparameters
\begin{equation}
    \begin{split}
        M_z &= [A_z, b_z]\\
        K_z &= \mathbbm 1_{n+1}.
    \end{split}
\end{equation}

\paragraph{\ac{sde} Dispersion}
We set 
\begin{equation}
\begin{split}
    \Psi_{D_z} &=  \Sigma_z * 0.1\quad \lambda_{D_z} = n + 2,
\end{split}
\end{equation}
with a heuristic downscaling as above.

\paragraph{Observation Covariance}
Lastly, 
\begin{equation}
\begin{split}
    \Psi_{x} &=  \Sigma_z * 0.5\quad \lambda_{D_z} = n + 2.
\end{split}
\end{equation}

\subsection{Inference of Gene-Switching Dynamics}
\begin{figure}
\centering
    \includegraphics[width=.6\columnwidth]{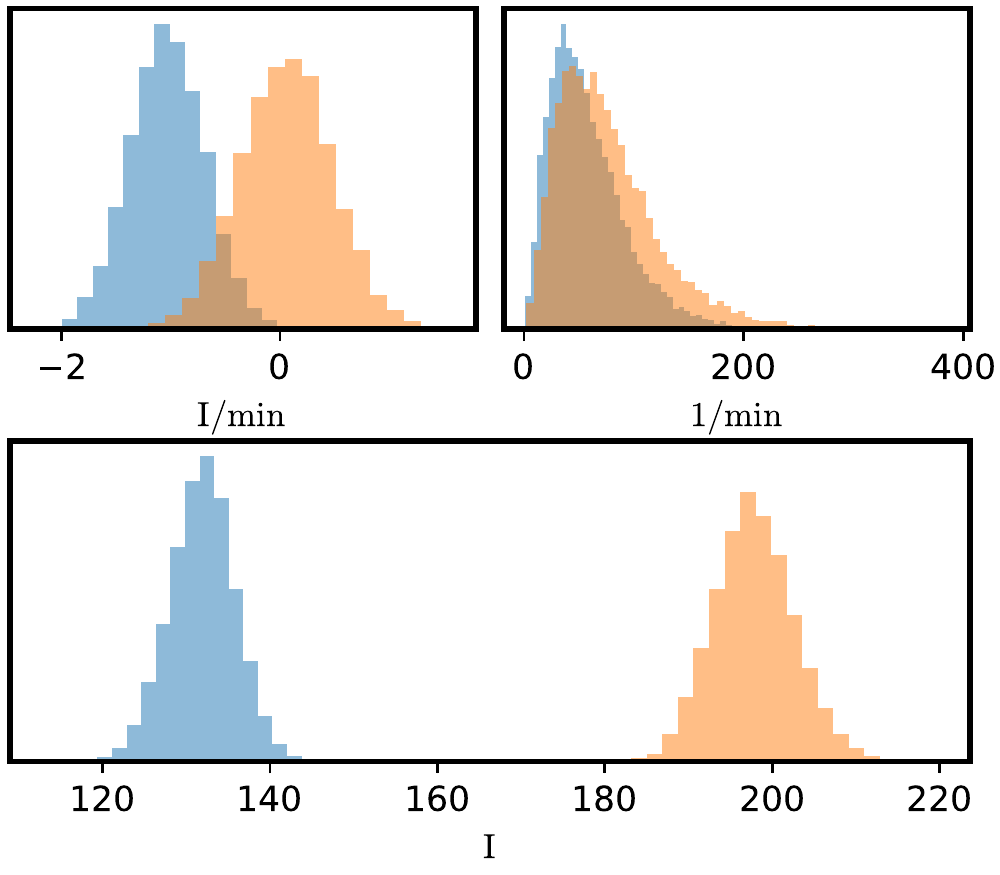}
    \caption{Bayesian parameter estimates. Top left: drift parameters $A_z$. Top right: rates $\Lambda_{zz'}$. Bottom: drift parameters $b_z$. $I$: fluorescence intensity (a.u.).}
    \label{fig:app_fluorescence_parameters}
\end{figure}

We provide the posterior parameter distributions in \cref{fig:app_fluorescence_parameters}.
We initialized the variational method up for comparison in the same way, but set the parameters directly instead of distribution hyperparameters, as this method does not take a fully Bayesian approach, but rather works with point estimates.


\end{document}